\documentclass[lettersize,journal]{IEEEtran}
\usepackage{amsmath,amsfonts}
\usepackage{array}
\usepackage[caption=false,font=normalsize,labelfont=sf,textfont=sf]{subfig}
\usepackage{textcomp}
\usepackage{stfloats}
\usepackage{url}
\usepackage{verbatim}
\usepackage{graphicx}

\usepackage{cite}
\usepackage{url}
\usepackage{colortbl}
\usepackage{hyperref}  
\usepackage{listings}
\usepackage{booktabs}
\usepackage{multirow}
\usepackage{upgreek}
\usepackage[ruled]{algorithm2e}

\usepackage[table]{xcolor}
\usepackage{adjustbox}
\usepackage{array}

\usepackage{pifont}

\newcommand{\cmark}{\ding{51}}
\newcommand{\xmark}{\ding{55}}

\newcommand{\subab}[1]{\hspace{0.9em}{\footnotesize$\triangleright$}~#1}

\usepackage[square, comma, sort&compress, numbers]{natbib}
\begin{document}
\title{EvoBrain: Continual Learning of EEG Foundation Models Across Heterogeneous BCI Tasks}

\author{Yangxuan Zhou*, Sha Zhao*, Jiquan Wang, Shijian Li, Gang Pan
	\thanks{* These authors contributed equally to this work.}
	\thanks{This work was supported by STI 2030 Major Projects (No. 2021ZD0200400), the Key Program of the Natural Science Foundattion of Zhejiang Province, China (No. Z24F020009) and Natural Science Foundation of China (No. 61925603). The corresponding author is Dr. Sha Zhao.}
	\thanks{Yangxuan Zhou, Sha Zhao, Jiquan Wang, Shijian Li and Gang Pan are with the State Key Laboratory of Brain-machine Intelligence and College of Computer Science and Technology, Zhejiang University, Hangzhou, Zhejiang, China. And Gang Pan is also with MOE Frontier Science Center for Brain Science and Brain-Machine Integration, Zhejiang University, Hangzhou, Zhejiang, China (e-mail: zyangxuan@zju.edu.cn; szhao@zju.edu.cn; wangjiquan@zju.edu.cn; shijianli@zju.edu.cn; gpan@zju.edu.cn).}
}


\maketitle

\begin{abstract}
Electroencephalography (EEG) serves as the predominant modality for non-invasive Brain-Computer Interfaces (BCIs) due to its high temporal resolution; however, conventional decoding approaches rely on fragmented, task-specific architectures that severely limit scalability across heterogeneous BCI tasks. Transcending these limitations, EEG foundation models pre-trained on massive corpora have emerged as a transformative solution for universal brain decoding, offering a unified framework that effectively addresses the generalization bottlenecks inherent in traditional machine learning pipelines. However, current post-training strategies rely on task-isolated fine-tuning. This static paradigm fundamentally restricts the effective transfer of knowledge across heterogeneous BCI tasks, thereby hindering the scalability of foundation models, while also incurring prohibitive computational and storage overheads that scale linearly with the number of tasks. In this paper, we formulate downstream adaptation of EEG foundation models as a cross-task continual learning problem and propose EvoBrain, a dynamic task-aware post-training continual learning framework that enables a unified EEG foundation model to generalize across diverse EEG tasks. EvoBrain addresses the plasticity-stability trade-off through two complementary components. Neuro-Spectral Task Normalization (NSN) aligns incoming tasks with historical task statistics while recalibrating task-specific spectral responses, enabling adaptation under both distributional and neuro-spectral shifts. Response-Affinity Distillation (RAD), combined with time-dependent replay, preserves old-task response geometry and promotes selective knowledge transfer between spectrally compatible tasks, mitigating forgetting without indiscriminate cross-task alignment. Extensive evaluations on six distinct BCI tasks demonstrate that our approach consistently surpasses state-of-the-art methods across diverse foundation backbones, effectively balancing the plasticity-stability trade-off. To the best of our knowledge, this work pioneers the exploration of cross-task continual learning in the EEG domain, advancing the realization of a unified one-for-all EEG decoding system.
\end{abstract}


\section{Introduction}

\IEEEPARstart{B}{rain-computer} interfaces (BCIs) provide a direct pathway for translating neural activity into control signals for external devices \cite{wu2016cyborg, schalk2004bci2000}. As the predominant non-invasive BCI technique, electroencephalography (EEG) records cortical electrical activity with high temporal resolution, enabling its widespread deployment across practical and clinical applications \cite{cowie2001emotion,jeong2004eeg,jenke2014feature}. Early studies typically relied on hand-crafted features combined with machine learning algorithms to analyze EEG signals \cite{lotte2007review}. With the advances in deep learning, a variety of models have been proposed to automatically decode EEG patterns and identify brain states across diverse experimental paradigms, including but not limited to sleep staging \cite{zhou2025personalized}, emotion recognition \cite{ding2022tsception}, and motor imagery \cite{altaheri2023deep}. 
These models are typically task-specific with different architectures, and are trained and evaluated on the same datasets, which inherently constrains their generalization capability and scalability across heterogeneous EEG tasks. 
To overcome this bottleneck, recent advancements have introduced EEG foundation models \cite{kostas2021bendr, yang2023biot, zhang2023brant, jiang2024large, wang2024eegpt, shi2024fome, jiang2024neurolm, wang2024cbramod, wang2025eegmamba, zhou2025csbrain, ding2025brainpro, chen2025uni, wang2026deeper}, which are pretrained in a self-supervised manner on large-scale unlabeled EEG data to learn universal, task-agnostic neural representations. In this manner, EEG foundation models can utilize a unified backbone architecture that requires only task-specific classifiers to be attached, and can then be effectively adapted to various downstream tasks via post-training (i.e., fine-tuning), achieving performance comparable to or even surpassing that of traditional task-specific models. 

\begin{figure}[!t]
	\centering
	\includegraphics[width=1.0\columnwidth]{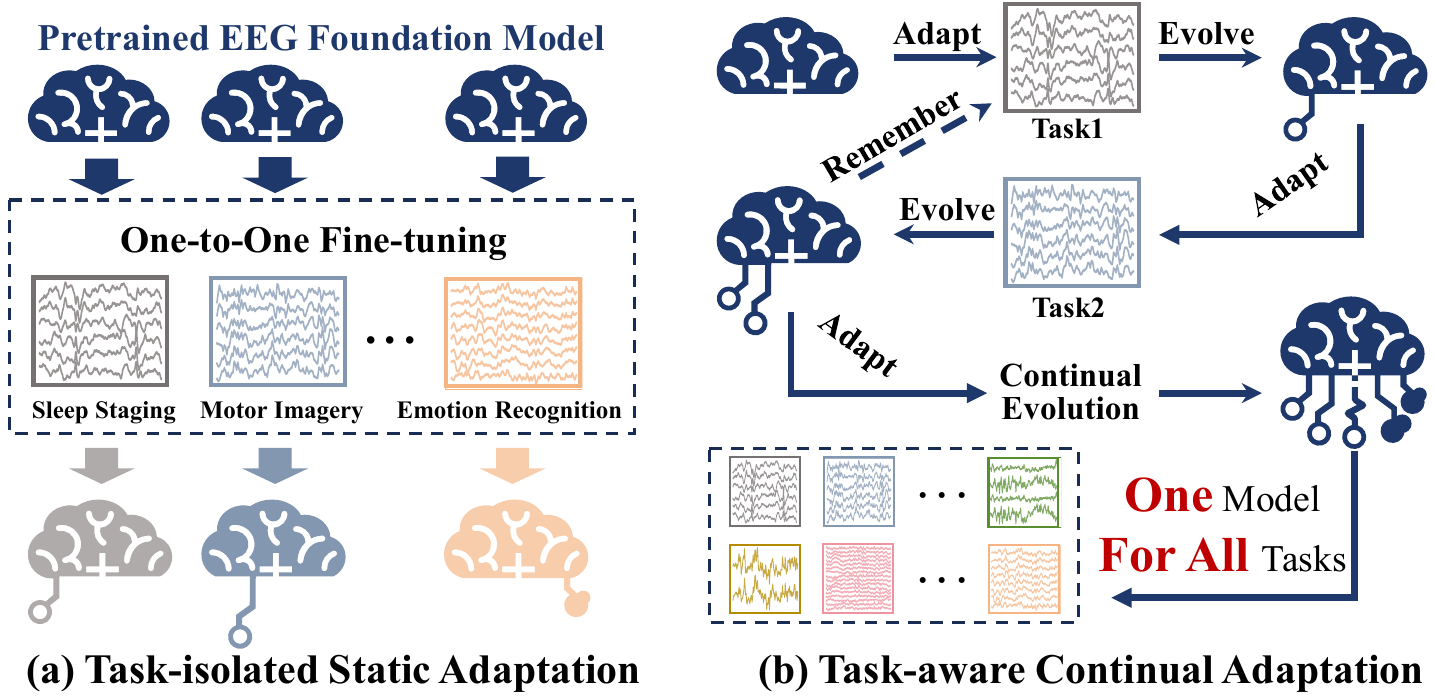}
	\caption{(a) The traditional task-isolated static post-training paradigm; (b) the task-aware continual adaptation strategy for a universal EEG decoding model.}
	\label{fig:intro}
\end{figure}
Despite the success of EEG foundation models, current research disproportionately prioritizes pre-training, leaving post-training adaptation largely underexplored. As illustrated in Fig.~\ref{fig:intro}(a), even with a unified backbone, existing methods predominantly resort to task-isolated full fine-tuning, a practice that imposes two fundamental limitations: 
\textbf{First, task-isolated adaptation inherently hinders cross-task knowledge transfer.} Unlike large language models (LLMs) that exhibit emergent abilities through scaling or cross-task learning  \cite{wei2022emergent, schaeffer2023emergent}, EEG foundation models under current post-training paradigms remain confined to narrow task boundaries. This isolation prevents the model from synthesizing higher-order, invariant neural patterns, thereby severely constraining its generalization potential beyond individual tasks. 
\textbf{Second, this static adaptation strategy struggles to accommodate the dynamic expansion requirements of EEG foundation models.} Adapting to a new EEG task necessitates a complete re-training from the pre-trained initialization and the storage of a separate set of parameters. Consequently, computational overhead and storage demands scale linearly with the number of tasks, creating prohibitive bottlenecks for scalable, multi-task deployment.
In summary, current static and task-isolated post-training strategies fundamentally hinder the realization of universal brain decoding models. There is an urgent need for a dynamic and task-aware post-training strategy that enables a single, unified EEG foundation model to generalize across diverse EEG tasks, ultimately achieving the one-for-all objective. Critically, this dynamic adaptation naturally formulates a Continual Learning (CL) problem \cite{wang2024comprehensive}, wherein the foundation model incrementally acquires knowledge from a sequence of heterogeneous EEG tasks without catastrophic forgetting. To succeed in this process, the framework must effectively maintain an appropriate trade-off between learning plasticity (i.e., the capacity to adapt to new tasks) and memory stability (i.e., the ability to retain knowledge on prior tasks.).

However, enabling a pretrained EEG foundation model to continually adapt to sequentially arriving different BCI tasks while preserving previously acquired capabilities remains highly challenging. Two coupled obstacles are central to this setting. First, learning plasticity is limited by the multi-level heterogeneity of EEG task streams. Different BCI paradigms may differ in acquisition settings, label semantics, signal morphology, and task-specific neuro-spectral patterns. Therefore, effective adaptation must go beyond generic distribution alignment to adapt task-specific temporal statistics and recalibrate task-relevant spectral responses. Second, memory stability is challenged by continual updates that may distort the response organization learned from previous tasks. In heterogeneous EEG streams, transferable structures are often only partially shared, and indiscriminate alignment across unrelated paradigms can induce negative transfer. Thus, continual adaptation must preserve old-task response relations while enabling selective transfer between tasks with compatible neural signatures.

To address these challenges, \textbf{we propose EvoBrain, a cross-task continual learning framework for adapting pretrained EEG foundation models toward one-for-all brain decoding.} EvoBrain is built upon two complementary principles. First, Neuro-Spectral Task Normalization (NSN) enhances learning plasticity by aligning incoming tasks with accumulated historical statistics while recalibrating task-specific spectral responses. Second, Response-Affinity Distillation (RAD), together with time-dependent replay, improves memory stability by preserving old-task response organization and promoting selective knowledge transfer between spectrally compatible tasks. In this way, EvoBrain enables a single EEG foundation backbone to evolve across heterogeneous BCI tasks without repeatedly restarting task-isolated fine-tuning. The framework is also plug-and-play, allowing integration with both Transformer-based and Mamba-based EEG foundation models without requiring backbone-specific architectural modifications. The key contributions of this paper are summarized as follows:

\begin{itemize}
	\item 
	\textbf{To the best of our knowledge, this work pioneers the exploration of cross-task continual learning in EEG-related analysis.} We propose a novel framework, termed EvoBrain, to overcome the critical limitation of static, task-isolated adaptation in pretrained EEG foundation models. The framework enables a single unified model to dynamically generalize across heterogeneous EEG tasks, thereby advancing toward the long-sought one-for-all objective for universal brain decoding. 
	
	\item 
	\textbf{We design Neuro-Spectral Task Normalization (NSN) and Response-Affinity Distillation (RAD) to address the plasticity-stability trade-off in cross-task continual EEG decoding.} NSN enhances adaptation to heterogeneous incoming tasks by jointly stabilizing temporal statistics and recalibrating task-specific spectral responses, while RAD preserves old-task response organization and promotes affinity-aware knowledge transfer across spectrally compatible tasks.
	
	\item
	\textbf{By integrating our framework with four mainstream EEG foundation models of diverse architectures, we demonstrate its architecture-agnostic efficacy}, achieving consistent improvements in both adaptation plasticity and memory stability across six EEG tasks.
\end{itemize}

\section{Related Works}
\subsection{EEG Decoding}
EEG employs scalp sensors to record the brain's spontaneous electrical activity. Early EEG decoding approaches predominantly relied on handcrafted features combined with traditional machine learning methods \cite{bashashati2007survey, lotte2007review}. These features included time-domain statistics, frequency-band power and spatial filters such as Common Spatial Patterns (CSP) \cite{wu2014probabilistic}. The extracted features were then fed into shallow classifiers, typically Linear Discriminant Analysis (LDA) \cite{balakrishnama1998linear} or Support Vector Machines (SVM) \cite{hearst1998support}. These methods achieved moderate performance but were sensitive to inter-subject variability and required extensive domain expertise for feature engineering. The advent of deep learning \cite{lecun2015deep} has shifted the paradigm toward end-to-end trainable architectures that automatically learn discriminative representations from raw or minimally preprocessed EEG signals. For example, Convolutional Neural Networks (CNNs), Recurrent Neural Networks (RNNs), and Long Short-Term Memory (LSTM) networks are widely used to capture the spatio-temporal dynamics inherent in EEG signals and have demonstrated superior accuracy across diverse experimental paradigms \cite{schirrmeister2017deep, sakhavi2018learning, wang2018lstm, abdelhameed2021deep}. The Transformer architecture has also been adopted to learn spatio-temporal features for various EEG decoding applications \cite{song2021transformer, du2022eeg, phan2022sleeptransformer}. To leverage the complementary advantages of CNNs and Transformers, several studies have proposed hybrid CNN–Transformer models for EEG classification tasks \cite{sun2021eeg, song2022eeg, xie2022transformer}. While these methods perform well on specific tasks or paradigm settings, their inherent task-specific architectural design limits their ability to generalize across diverse EEG tasks, thereby restricting their applications in practice.

\subsection{EEG Foundation Models}
Inspired by the remarkable success of foundation models in natural language processing \cite{achiam2023gpt, devlin2019bert}, computer vision \cite{he2022masked, kirillov2023segment}, and multimodal perception \cite{radford2021learning}, a growing body of recent work has begun to explore their application to the decoding of brain signals. EEG foundation models typically follow a pretraining–post-training paradigm \cite{lai2025simple, kuruppu2025eeg, xiong2025eeg}: they are first pretrained on large-scale unlabeled EEG data and subsequently adapted to diverse downstream tasks through post-training strategies. Current research has primarily focused on improving the learning of general-purpose EEG representations during pretraining, while largely neglecting the design of effective post-training adaptation mechanisms. Existing adapting approaches can be broadly categorized into two types: (i) single-task fine-tuning \cite{zhang2023brant, zhang2024brant, jiang2024large, wang2024eegpt, shi2024fome, chen2024eegformer, cui2024neuro, wang2024cbramod, wang2025eegmamba, zhou2025csbrain, ding2025brainpro, ma2025codebrain, li2025comet}, which fine-tunes the entire model separately for each task, leading to linear growth in computational and storage costs with respect to the number of tasks; and (ii) multi-task fine-tuning \cite{daimteeg, jiang2024neurolm, chen2025uni, lu2025unimind}, which shares backbone parameters across tasks during joint optimization. Although the latter mitigates resource overhead, it remains a static adaptation scheme that requires re-fine-tuning whenever a new task is introduced and employs only naive multi-task learning without explicitly modeling cross-task knowledge transfer. Consequently, there exists a compelling need for a dynamic and task-aware adaptation mechanism to overcome the inherent limitations of current post-training strategies for EEG foundation models.


\subsection{Continual EEG Decoding}
Studies in continual learning can be broadly categorized into three major paradigms: regularization-based methods \cite{kirkpatrick2017overcoming, zenke2017continual}, parameter isolation–based methods \cite{rusu2016progressive, mallya2018packnet}, and rehearsal-based methods \cite{rebuffi2017icarl, castro2018end}. Recently, research has increasingly focused on continual EEG decoding, a setting well-suited to real-world scenarios wherein static decoding models are unable to handle the continual and dynamic emergence of new subjects and tasks. Existing efforts, however, are predominantly confined to a single EEG paradigm, devising individual-level continual decoders to mitigate the continual domain shift caused by inter-subject variability \cite{duan2023replay, shahbazinia2024resource, jin2024affective, li2025personalized, zhoubrainuicl, zhou2026spiced}. To the best of our knowledge, the exploration of EEG-based task-level continual learning remains entirely open. This presents a substantially greater challenge, as the macroscopic disparities between distinct tasks are far more pronounced than the microscopic differences among individuals. To address this, we propose EvoBrain, a dynamic and task-aware continual decoding framework. EvoBrain enables a single, unified EEG foundation model to incrementally acquire and consolidate cross-task knowledge, ultimately evolving into a general-purpose expert for EEG understanding.

\section{Methodology}
\subsection{Problem Setup and Preliminaries}
This work aim to enable a pretrained EEG foundation model to continually adapt to a sequence of heterogeneous EEG decoding tasks while preserving its performance on previously encountered ones. We formalize this setting as cross-task continual learning. Formally, let $\{\mathcal{T}_1, \mathcal{T}_2, \mathcal{T}_3, \dots, \mathcal{T}_\mathcal{N}\}$ denote a stream of $\mathcal{N}$ distinct EEG tasks (e.g., sleep staging, motor imagery, emotion recognition). Each task $\mathcal{T}_i$ is associated with a unique data distribution $\mathcal{D}_i$, where $\mathcal{D}_i \neq \mathcal{D}_j$ and $ 1\leq i \neq j \leq \mathcal{N}$, reflecting substantial domain shifts in signal statistics and label spaces across tasks. Let $\mathcal{M}$ denotes the continual learning model, where $\mathcal{M}_0$ represents the initial pretrained EEG foundation model and $\mathcal{M}_i$ denotes the updated model after adaptation to the $i$-th task $\mathcal{T}_i$. Our core objective is to simultaneously optimize two competing criteria, as formally described below:

(1) \textbf{Learning plasticity:} The continual learning model $\mathcal{M}_i$ should be able to adapt effectively to a new task $\mathcal{T}_i$.
\begin{equation}\label{eq1}
	\min_{\theta_\mathcal{M}} (\mathbb{E}_{(x, y) \sim \mathcal{D}_i}  \mathcal{L}(\mathcal{M}_i(x), y))
\end{equation}
where $\mathcal{M}$ parameterized by $\theta_\mathcal{M}$ and $\mathcal{L}$ denotes a task-specific loss that measures the prediction error of the model $\mathcal{M}_i$ on the given task $\mathcal{T}_i$.

(2) \textbf{Memory stability:} The updated model $\mathcal{M}_i$ should retain performance on all previously learned tasks $\{\mathcal{T}_1, \mathcal{T}_2, \dots, \mathcal{T}_{j}\}$, i.e., for all $j < i$.
\begin{equation}\label{eq2}
\mathbb{E}_{(x,y)\sim\mathcal{D}_j} \left[ \mathcal{L}(\mathcal{M}_j(x), y) \right] \approx \mathbb{E}_{(x,y)\sim\mathcal{D}_j} \left[ \mathcal{L}(\mathcal{M}_i(x), y) \right]
\end{equation}
The fundamental challenge lies in mitigating catastrophic forgetting, the tendency of $\mathcal{M}_i$ to overfit to $\mathcal{D}_i$ and degrade performance on $\mathcal{D}_j$ ($j < i$ ), while effectively acquiring new knowledge from $\mathcal{D}_i$. This necessitates a dynamic trade-off between plasticity for adaptation and stability for memory retention. To this end, we formulate the continual learning objective under task $\mathcal{T}_i$ as follows:
\begin{equation}\label{eq3}
	\small
	\min_{\mathcal{M}_i} \underbrace{\mathbb{E}_{(x,y)\sim\mathcal{D}_i} \mathcal{L}(\mathcal{M}_i(x), y)}_{\text{New Task adaptation}} +  \underbrace{\mathcal{R}\big(\mathcal{M}_i; \{\mathcal{M}_j\}_{j=1}^{i-1}, \{\mathcal{D}_j\}_{j=1}^{i-1}\big)}_{\text{Historical knowledge retention}}
\end{equation}
where $\mathcal{R}(\cdot)$ is a generic regularization term that penalizes deviations from the knowledge encoded in previously learned models $\{\mathcal{M}_j\}_{j=1}^{i-1}$ with respect to their corresponding data distributions $\{\mathcal{D}_j\}_{j=1}^{i-1}$. Our goal is to learn a universal EEG decoder $\mathcal{M}_\mathcal{N}$ that generalizes across all learned $\mathcal{N}$ tasks without requiring retraining, thus realizing the one model for all tasks vision.

\subsection{Overview}
\begin{figure*}[!t]
	\centering
	\includegraphics[width=1.0\textwidth]{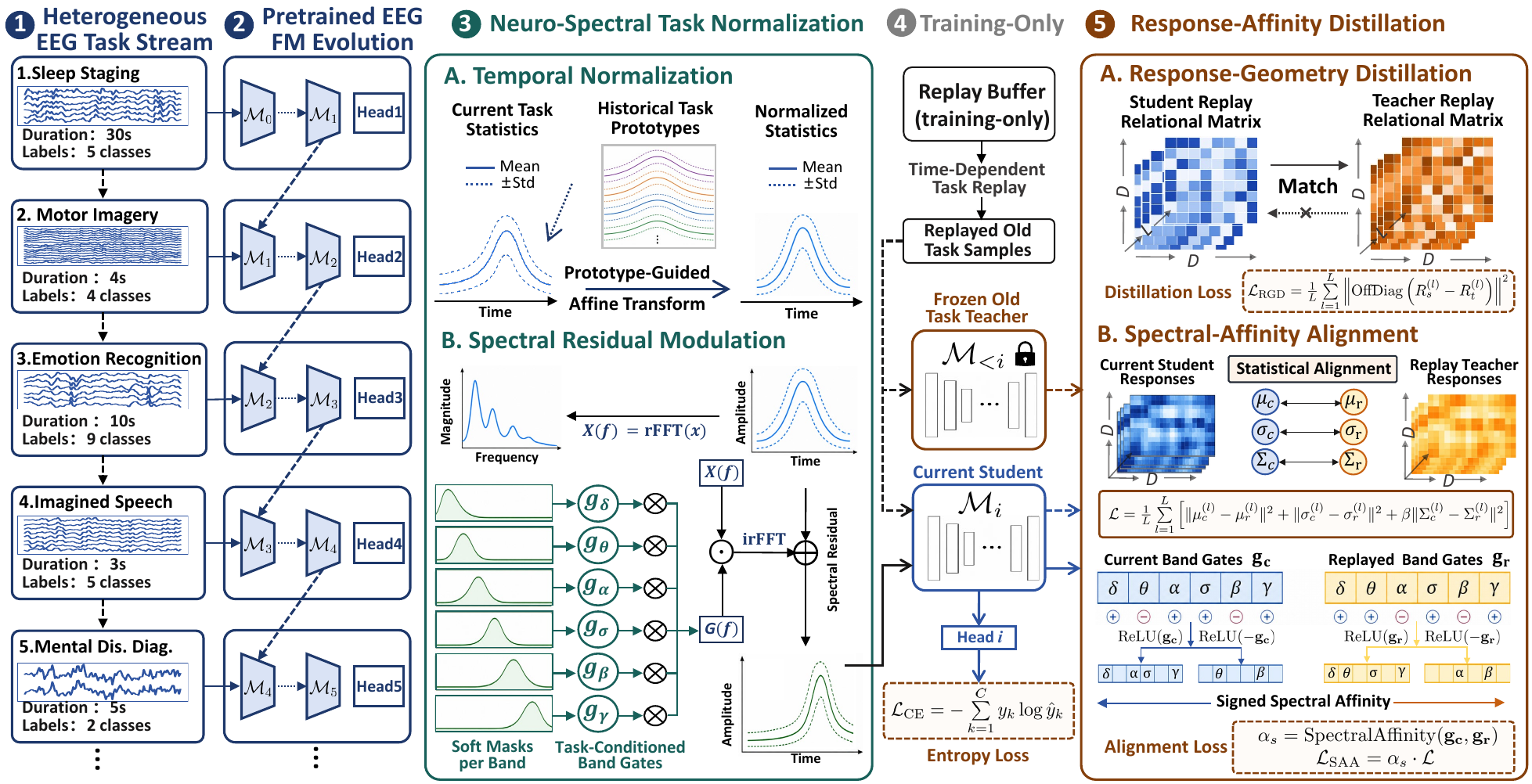}
	\caption{\textbf{Overview of EvoBrain for continual learning of EEG foundation models across heterogeneous BCI tasks.}
		(1) EvoBrain receives a stream of heterogeneous EEG tasks with different signal durations, channels and label spaces.
		(2) Starting from a pretrained EEG foundation model, the backbone is progressively adapted from $\mathcal{M}_0$ to $\mathcal{M}_1,\ldots,\mathcal{M}_\mathcal{N}$, while lightweight task-specific heads are attached for each task.
		(3) For each incoming task, Neuro-Spectral Task Normalization (NSN) first performs prototype-guided temporal normalization to reduce cross-task distribution shifts, and then applies task-conditioned spectral residual modulation by combining band-specific soft masks and learnable frequency-band gates.
		(4) During training, replayed old-task samples and current new-task samples are jointly processed by the frozen old-task teacher and the current student model; dashed arrows denote replayed old-task samples, whereas solid arrows denote current new-task samples.
		(5) Response-Affinity Distillation (RAD) regularizes continual adaptation with two complementary objectives: Response-Geometry Distillation preserves the relational geometry of old-task responses between the student and teacher, and Spectral-Affinity Alignment aligns cross-task response statistics according to the signed similarity of task-conditioned spectral band gates.
		At inference time, EvoBrain only requires the continually adapted backbone and the selected task head.
	}
	\label{fig:framework}
\end{figure*}
EvoBrain is a continual adaptation framework that evolves a pretrained EEG foundation model across a stream of heterogeneous BCI tasks, as illustrated in Fig.~\ref{fig:framework}. Starting from a pretrained EEG foundation model $\mathcal{M}_0$, EvoBrain sequentially updates the backbone from $\mathcal{M}_{i-1}$ to $\mathcal{M}_i$ as new heterogeneous tasks arrive, while attaching a lightweight task-specific classifier head for each task. Upon the arrival of a new task $\mathcal{T}_i$, EvoBrain first applies Neuro-Spectral Task Normalization (NSN), which performs historical prototype-guided temporal normalization to reduce cross-task distribution shifts and further introduces spectral residual modulation to adapt task-specific neural responses across standard EEG frequency bands. Then, following the rehearsal principle in continual learning \cite{rebuffi2017icarl, castro2018end}, we adopt a time-dependent replay strategy and introduces Response-Affinity Distillation (RAD) to regularize the current adaptation. Specifically, RAD preserves previous-task stability through response-geometry distillation on replayed samples and promotes transferable high-order neural representations through spectral-affinity response alignment across tasks. During test stage, EvoBrain only retains the continually adapted backbone and the corresponding task-specific head for inference. Notably, EvoBrain can be integrated into mainstream EEG foundation models, including both Transformer-based and Mamba-based architectures, without requiring complex architectural modifications.

\subsection{Time-Dependent Task Replay}

EvoBrain employs a lightweight time-dependent replay strategy to maintain historical task coverage during sequential adaptation. Since earlier tasks undergo more subsequent updates and are typically more vulnerable to forgetting, we assign larger replay probabilities to older tasks instead of sampling all stored tasks uniformly \cite{castro2018end}. 
Let $\{\mathcal{T}_1,\mathcal{T}_2,\dots,\mathcal{T}_{i-1}\}$ be the historical tasks before learning $\mathcal{T}_i$. 
For each task $\mathcal{T}_t$, its temporal depth and replay weight are defined as
\begin{equation}\label{replay}
	d_t = i - t + 1, 
	\qquad
	w_t = \left( 1 - e^{-0.8 \cdot d_t} \right)
\end{equation}
In practice, each current-task mini-batch is paired with one replay mini-batch during training. We sample one historical task according to the replay probabilities $\{w_t\}_{t=1}^{i-1}$, and then draw a same-sized mini-batch from its stored training samples. The replay batch only provides old-task responses for distillation, without introducing an additional retrieval network or replay optimization module. Thus, the training-time replay cost is limited to one auxiliary historical mini-batch per iteration, and no replay is required at inference.

\subsection{Neuro-Spectral Task Normalization}
\label{sec:nsn}

Heterogeneous EEG tasks differ substantially in signal morphology, spatial topography, and label semantics, causing severe inter-task domain shifts during sequential adaptation. Beyond these global distributional discrepancies, different BCI paradigms often rely on distinct neuro-spectral response patterns over standard EEG frequency bands.  When a pretrained EEG foundation model is exposed to such non-stationary task streams, both data statistics and task-related spectral responses may drift across tasks, impairing learning plasticity and memory stability.  To address this, we propose \emph{Neuro-Spectral Task Normalization} (NSN), which integrates Prototype-guided Temporal Normalization (PTN) to stabilize cross-task statistics and task-conditioned Spectral Residual Modulation (SRM) to adaptively recalibrate task-relevant frequency responses.

\textbf{Prototype-guided Temporal Normalization:}
Given an input tensor $\mathbf{X} \in \mathbb{R}^{\mathbf{B} \times \mathbf{C} \times \mathbf{D} \times d}$ from the current task $\mathcal{T}_i$, $\mathbf{B}$, $\mathbf{C}$, and $\mathbf{D}$ denote the batch size, number of channels, and duration of EEG segments, respectively, while $d$ denotes the fixed patch embedding dimension of the EEG foundation model. For each task $\mathcal{T}_i$, we maintain a historical task prototype $\mathbf{p}_i \in \mathbb{R}^{d}$, which is updated as an exponential moving average of the task-wise data mean of $\mathbf{X}$ over the batch, channel, and duration dimensions during the first training epoch. When a new input arrives, we compute the current-task statistics $(\mu_i,\sigma_i)$ from $\mathbf{X}$ and derive the historical reference statistics $(\mu_g,\sigma_g)$ from the set of previous task prototypes $\mathcal{P}_{<i}=\{\mathbf{p}_1,\ldots,\mathbf{p}_{i-1}\}$. Here, $(\mu_g,\sigma_g)$ summarize the historical feature distribution accumulated before learning $\mathcal{T}_i$.

Instead of forcing the current task $\mathcal{T}_i$ to rigidly match the historical statistics of $\mathcal{T}_{<i}$, we utilize learnable task-specific moment-gating parameters $\eta_{\mu}^{i}, \eta_{\sigma}^{i} \in \mathbb{R}$ to dynamically rescale and shift the $i$-th task data. We define gating coefficients as follows:
\begin{equation}
	\lambda_{\mu}^{i} = \sigma(\eta_{\mu}^{i}), 
	\qquad
	\lambda_{\sigma}^{i} = \sigma(\eta_{\sigma}^{i}),
\end{equation}
where $\sigma(\cdot)$ denotes the sigmoid function. The target moments are then obtained by
\begin{align}
	\mu_i^{\mathrm{tar}} 
	&= \lambda_{\mu}^{i}\mu_g + (1-\lambda_{\mu}^{i})\mu_i, \\
	\sigma_i^{\mathrm{tar}} 
	&= \lambda_{\sigma}^{i}\sigma_g + (1-\lambda_{\sigma}^{i})\sigma_i .
\end{align}
The affine transformation parameters are then:
\begin{align}
	\gamma_i &= \frac{\sigma_i^{\mathrm{tar}}}{\sigma_i}, \\
	\beta_i  &= \mu_i^{\mathrm{tar}} - \gamma_i \mu_i,
\end{align}
and normalizes the current task input $\mathbf{X}$ as
\begin{equation}
	\mathbf{X}_{\mathbf{tn}} = \gamma_i \mathbf{X} + \beta_i .
\end{equation}

\textbf{Task-conditioned Spectral Residual Modulation:}
After temporal normalization, we further introduce spectral residual modulation to explicitly model task-specific spectral response patterns. Given the temporally normalized representation $\mathbf{X}_{\mathrm{tn}}$, we first compute its real-valued Fourier spectrum along the within-patch temporal dimension:
\begin{equation}
	\widehat{\mathbf{X}}(f) = \operatorname{rFFT}(\mathbf{X}_{\mathbf{tn}}).
\end{equation}

Let $\mathcal{B}$ denote the standard EEG bands: $\delta$ $(0.5$--$4~\mathrm{Hz})$, $\theta$ $(4$--$8~\mathrm{Hz})$, $\alpha$ $(8$--$13~\mathrm{Hz})$, $\sigma$ $(12$--$16~\mathrm{Hz})$, $\beta$ $(16$--$30~\mathrm{Hz})$, and $\gamma$ $(30$--$45~\mathrm{Hz})$. For each band $b \in \mathcal{B}$ with lower and upper boundaries $(l_b,u_b)$, we construct a soft band mask $M_b(f)$ with a smooth transition width $\tau=1$ Hz:
\begin{equation}
	M_b(f)=
	\begin{cases}
		1, & l_b \leq f \leq u_b, \\[2pt]
		\frac{1}{2}-\frac{1}{2}\cos\!\left(\pi \frac{f-(l_b-\tau)}{\tau}\right), 
		& l_b-\tau \leq f < l_b, \\[4pt]
		\frac{1}{2}+\frac{1}{2}\cos\!\left(\pi \frac{f-u_b}{\tau}\right), 
		& u_b < f \leq u_b+\tau, \\[4pt]
		0, & \mathrm{otherwise}.
	\end{cases}
\end{equation}
This construction assigns full weights to frequencies inside each EEG band and smoothly attenuates frequencies near band boundaries. 
To handle overlapping frequency regions, the masks are normalized across bands as follows:
\begin{equation}
	W_b(f) =
	\frac{M_b(f)}
	{\sum_{b' \in \mathcal{B}} M_{b'}(f)+ \epsilon}
\end{equation}
where $\epsilon$ is a small positive constant for numerical stability. For the current task, we assign a learnable parameter $a_b$ to each band $b \in \mathcal{B}$ and obtain the corresponding band gate by
\begin{equation}
	g_b = \tanh(a_b), \qquad b \in \mathcal{B},
\end{equation}
where $g_b \in [-1,1]$ controls whether the response of band $b$ is enhanced or suppressed. The band-level gates are then projected to frequency bins as
\begin{equation}
	G(f) = \sum_{b \in \mathcal{B}} g_b W_b(f),
\end{equation}
yielding a task-conditioned spectral modulation function. 
The residual spectrum is computed by
\begin{equation}
	\Delta \widehat{\mathbf{X}}(f) = G(f) \odot \widehat{\mathbf{X}}(f),
\end{equation}
and transformed back to the feature domain:
\begin{equation}
	\Delta \mathbf{X} = \operatorname{irFFT}\big(\Delta \widehat{\mathbf{X}}(f)\big).
\end{equation}
Finally, NSN applies residual modulation:
\begin{equation}\label{eq_nsn}
	\mathbf{X}_{\mathrm{nsn}}
	=
	\mathbf{X}_{\mathbf{tn}} + \alpha_m \Delta \mathbf{X},
\end{equation}
where $\alpha_m$ controls the modulation strength. 

To quantify the frequency-wise effect induced by this residual modulation, we further define a relative linear power difference:
\begin{equation}\label{eq_relative_power_diff}
	\Delta_{\mathrm{rel}}(f)=
	\frac{P_{\mathrm{NSN}}(f)-P_{\mathrm{PTN}}(f)}
	{\frac{1}{|\Omega|}\sum_{f'\in\Omega}P_{\mathrm{PTN}}(f')},
	\quad \Omega=[0.5,45]\mathrm{Hz},
\end{equation}
where
\(
P_{\mathrm{PTN}}(f)=|\operatorname{rFFT}(\mathbf{X}_{\mathrm{tn}})(f)|^2
\)
and
\(
P_{\mathrm{NSN}}(f)=|\operatorname{rFFT}(\mathbf{X}_{\mathrm{nsn}})(f)|^2
\).
This quantity measures the normalized power change from the temporally normalized representation to the NSN-modulated representation. A positive value of $\Delta_{\mathrm{rel}}(f)$ indicates that SRM amplifies the response around frequency $f$, whereas a negative value indicates suppression. Therefore, $\Delta_{\mathrm{rel}}(f)$ provides an interpretable frequency-wise view of how the learned task-conditioned gates reshape spectral responses.

By preserving the temporally normalized representation, this residual formulation enables task-adaptive enhancement or suppression of relevant spectral responses. Meanwhile, the learned band gate $g_b$ serves as a compact task-specific spectral signature for subsequent cross-task affinity estimation.

\subsection{Response-Affinity Distillation}

After Neuro-Spectral Task Normalization, the current EEG input is temporally aligned and spectrally recalibrated with task-conditioned band responses. 
However, sequential adaptation still faces two coupled challenges: preserving the response structure learned from previous tasks and extracting transferable high-order representations from heterogeneous tasks. 
To address this, we propose \emph{Response-Affinity Distillation} (RAD), which regularizes continual adaptation through two complementary objectives: Response-Geometry Distillation (RGD) for old-task stability and Spectral-Affinity Alignment (SAA) for cross-task transfer. When learning a new task $\mathcal{T}_i$, EvoBrain samples replay data from a previous task $\mathcal{T}_j$ $(j<i)$ using the time-dependent replay strategy and loads the corresponding frozen checkpoint $\mathcal{M}_j$, saved immediately after learning $\mathcal{T}_j$, as the teacher. The replayed samples are fed into both the current student model $\mathcal{M}_i$ and the frozen teacher model $\mathcal{M}_{<i}$. For the $\ell$-th encoder layer, we denote their intermediate responses as $\mathbf{H}_{s,r}^{(\ell)}$ and $\mathbf{H}_{t,r}^{(\ell)}$, respectively, where the subscripts $s$, $t$, and $r$ indicate student, teacher, and replayed old-task samples. Similarly, the response of current-task samples in the student model is denoted as $\mathbf{H}_{s,c}^{(\ell)}$. We then apply global average pooling over the channel and duration dimensions to these intermediate responses, yielding scale-consistent representations in $\mathbb{R}^{\mathbf{B} \times d}$ across heterogeneous EEG tasks for subsequent relation and statistic computation.

\textbf{Response-Geometry Distillation:}
RGD preserves old-task stability by matching the response geometry of replayed samples between the current student and the frozen old-task teacher. 
For the $\ell$-th encoder layer, we first $\ell_2$-normalize the pooled replay responses and compute their sample-wise relational matrices:
\begin{align}\label{eq19}
	\mathbf{R}_{s,r}^{(\ell)}
	&=
	\bar{\mathbf{H}}_{s,r}^{(\ell)}
	\bar{\mathbf{H}}_{s,r}^{(\ell)\top}, \\
	\qquad
	\mathbf{R}_{t,r}^{(\ell)}\label{eq20}
	&=
	\bar{\mathbf{H}}_{t,r}^{(\ell)}
	\bar{\mathbf{H}}_{t,r}^{(\ell)\top},
\end{align}
where $\bar{\mathbf{H}}$ denotes the $\ell_2$-normalized response representation. RGD matches only the off-diagonal entries of the two relational matrices:
\begin{equation}\label{eq21}
	\mathcal{L}_{\mathrm{RGD}}
	=
	\frac{1}{L}
	\sum_{\ell=1}^{L}
	\frac{1}{\mathbf{B}(\mathbf{B}-1)}
	\sum_{i=1}^{\mathbf{B}}
	\sum_{\substack{j=1 \\ j\neq i}}^{\mathbf{B}}
	\left(
	R_{s,r,ij}^{(\ell)}
	-
	R_{t,r,ij}^{(\ell)}
	\right)^2 .
\end{equation}
where $L$ denotes the number of encoder layers, $\mathbf{B}$ denotes the batch size, and $R_{s,r,ij}^{(\ell)}$ and $R_{t,r,ij}^{(\ell)}$ denote the $(i,j)$-th entries of the student and teacher relational matrices at layer $\ell$, respectively. The diagonal entries are excluded because they only measure self-similarities, whereas the off-diagonal entries characterize the pairwise response geometry among replayed samples. In this way, the proposed Response-Geometry Distillation preserves the old-task response geometry encoded by the frozen teacher, i.e., the past model immediately after adaptation to the replayed task, and retains relational knowledge to mitigate catastrophic forgetting during new-task adaptation.

\textbf{Spectral-Affinity Alignment:}
RGD focuses on old-task stability, whereas SAA encourages transferable response alignment between current and previous tasks. For each layer, we compute the response statistics of current-task student responses $\mathbf{H}_{s,c}^{(\ell)}$ and replay-teacher responses $\mathbf{H}_{t,r}^{(\ell)}$, denoted as $(\boldsymbol{\mu}_{c}^{(\ell)},\boldsymbol{\sigma}_{c}^{(\ell)},\boldsymbol{\Sigma}_{c}^{(\ell)})$ and $(\boldsymbol{\mu}_{r}^{(\ell)},\boldsymbol{\sigma}_{r}^{(\ell)},\boldsymbol{\Sigma}_{r}^{(\ell)})$, respectively. The statistical alignment loss is defined as follows:
\begin{equation}
	\begin{aligned}
		\mathcal{L}_{\mathrm{align}}
		=
		\frac{1}{L}
		\sum_{\ell=1}^{L}
		\Big(
		&\|\boldsymbol{\mu}_{c}^{(\ell)}
		-\boldsymbol{\mu}_{r}^{(\ell)}\|_{2}^{2}
		+
		\|\boldsymbol{\sigma}_{c}^{(\ell)}
		-\boldsymbol{\sigma}_{r}^{(\ell)}\|_{2}^{2} \\
		&+
		\rho
		\|\boldsymbol{\Sigma}_{c}^{(\ell)}
		-\boldsymbol{\Sigma}_{r}^{(\ell)}\|_{F}^{2}
		\Big).
	\end{aligned}
\end{equation}
where $\rho$ balances covariance alignment.

To avoid indiscriminate alignment between unrelated EEG paradigms, we weight this loss by cross-task spectral affinity. Specifically, let $\mathbf{g}_{c}=[g_{c,b}]_{b\in\mathcal{B}}$ and $\mathbf{g}_{r}=[g_{r,b}]_{b\in\mathcal{B}}$ denote the NSN band-gate vectors of the current and replayed tasks, respectively.
We estimate their signed spectral affinity $\alpha_s$ by comparing both positive and negative gate vectors as follows:
\begin{align}
	s_{\mathrm{pos}}
	&=
	\cos\big(\operatorname{ReLU}(\mathbf{g}_{c}),
	\operatorname{ReLU}(\mathbf{g}_{r})\big), \\
	s_{\mathrm{neg}}
	&=
	\cos\big(\operatorname{ReLU}(-\mathbf{g}_{c}),
	\operatorname{ReLU}(-\mathbf{g}_{r})\big), \\
	\alpha_s
	&=
	\operatorname{ReLU}
	\left(
	\frac{s_{\mathrm{pos}}+s_{\mathrm{neg}}}{2}
	\right).
\end{align}
The spectral-affinity response alignment loss is then
\begin{equation}
	\mathcal{L}_{\mathrm{SAA}}
	=
	\alpha_s \cdot \mathcal{L}_{\mathrm{align}}.
\end{equation}
Thus, response statistics are aligned only when the current and replayed tasks exhibit similar task-conditioned spectral signatures, reducing the risk of negative transfer across heterogeneous BCI paradigms. The overall RAD objective is
\begin{equation}
	\mathcal{L}_{\mathrm{RAD}}
	=
	\lambda_r \mathcal{L}_{\mathrm{RGD}}
	+
	(1-\lambda_{r})
	\mathcal{L}_{\mathrm{SAA}}
\end{equation}
where $\lambda_{r}$ controls the trade-off between old-task response-geometry retain and spectral-affinity-guided cross-task response alignment.
\begin{algorithm}[t] 
	\caption{Procedure of Cross-Task Continual Learning for Pretrained EEG Foundation model}
	\label{algorithm1}
	\KwIn{Sequential EEG tasks $\{\mathcal{T}_i\}_{i=1}^\mathcal{N}$, pretrained EEG foundation model $\mathcal{E}$, paradigm storage $\mathcal{M}$}
	
	\KwOut{EEG foundation Model $\mathcal{E}$, set of task-specific classifiers $\{\mathcal{C}_i\}_{i=1}^\mathcal{N}$}
	\textbf{Training Stage:}
	
	$\mathcal{M} \gets \emptyset$\;
	\For{$i\leftarrow1$ to $\mathcal{N}$}{
		\eIf{$i = 1$}{
			Optimize $\mathcal{E}$ and $\mathcal{C}_1$ by minimizing Eq. \ref{loss_ce}\;
			Store task $\mathcal{T}_1$ into $\mathcal{M}$\;
		}{
			Replay historical tasks from $\mathcal{M}$ by Eq. \ref{replay}\;
			Normalize input data via NSN by Eq. \ref{eq_nsn}\;
			Optimize $\mathcal{E}$ and $\mathcal{C}_i$ by Eq. \ref{total_loss}\;
			Store task $\mathcal{T}_i$ into $\mathcal{M}$\;
		}
	}
	\textbf{Test Stage:} 
	
	Normalize test data by Eq. \ref{eq_nsn} and predict with $\mathcal{C}_i$\;
\end{algorithm}
\subsection{Overall Loss Function}
\label{sec:overall_loss}

For the current task $\mathcal{T}_i$, we optimizes the task-specific classifier head with a supervised classification loss. For multi-class classification tasks, we use the categorical cross-entropy loss, while for binary classification tasks, we use the binary cross-entropy loss with logits. We denote this task-adaptation loss as $\mathcal{L}_\mathrm{CE}$. For $i>1$, EvoBrain further incorporates the replay-based RAD loss to preserve old-task knowledge and promote cross-task transfer. The overall training objective is defined as follows:
\begin{equation}\label{total_loss}
	\mathcal{L}
	=
	\lambda_{ce}\mathcal{L}_\mathrm{CE}
	+
	(1-\lambda_{ce})\mathcal{L}_{\mathrm{RAD}},
\end{equation}
where $\lambda_{ce}$ controls the trade-off between plasticity for the current task and stability for previously learned tasks. For the first incoming task $\mathcal{T}_1$, no historical task is available for replay or distillation; therefore, the objective reduces to
\begin{equation}\label{loss_ce}
	\mathcal{L}
	=
	\mathcal{L}_\mathrm{CE}.
\end{equation}

Overall, NSN and RAD play complementary roles in EvoBrain. NSN normalizes the incoming EEG representation by stabilizing temporal statistics and recalibrating task-conditioned spectral responses, while RAD regularizes the adapted model by preserving old-task response geometry and aligning transferable cross-task response statistics according to spectral affinity.  Together, they enable the pretrained EEG foundation model to continually adapt to heterogeneous BCI tasks while balancing new-task plasticity and old-task stability. The detailed algorithmic pipeline is illustrated in Algorithm \ref{algorithm1}.

\section{Experimental Setup}
\subsection{Datasets and Tasks}
\begin{table*}[t]
	\centering 
	\caption{Overview of downstream BCI tasks and datasets.}
	\resizebox{0.9\textwidth}{!}{\begin{tabular}{lccccccc}
		\toprule[1pt]
		BCI Tasks                                         & Datasets     & Rate  & \# Channels & Duration & \# Samples  & \# Subjects & Label \\ \midrule
		I. Emotion Recognition                            & FACED \cite{chen2023large}       & 250Hz & 32          & 10s                          & 10,332       & 123                  & 9-class                  \\
		II. Motor Imagery Classification & PhysioNet-MI \cite{schalk2004bci2000} & 160Hz & 64          & 4s                           & 9,837            & 109              & 4-class                   \\
		& BCIC-IV-2a \cite{brunner2008bci}  & 250Hz & 22          & 4s                           & 5088                & 9           & 4-class                   \\
		III. Sleep Staging                                & ISRUC \cite{isruc}        & 200Hz & 6           & 30s                          & 89,240        & 100                 & 5-class                   \\
		IV. Imagined Speech Classification                 & BCIC2020-3 \cite{jeong20222020}  & 256Hz & 64          & 3s                           & 6,000         & 15                 & 5-class                   \\
		V. Mental Disorder Diagnosis                      & Mumtaz2016 \cite{mumtaz2017electroencephalogram}  & 256Hz & 19          & 5s                           & 7,143     & 34                     & 2-class                 \\ \bottomrule[1pt] 
	\end{tabular}}
\label{tab1}
\end{table*}
To validate the efficiency of our EvoBrain framework, we systematically evaluate its decoding performance in a cross-task continual learning scenario across six downstream EEG tasks, as shown in Table \ref{tab1}.
\subsubsection{FACED Dataset}
This is a large-scale, high-resolution affective computing dataset that captures nine distinct affective states: amusement, inspiration, joy, tenderness, anger, fear, disgust, sadness, and neutral. The dataset includes EEG recordings from 123 participants, acquired using a 32-channel system at a sampling rate of 250 Hz. The continuous signals were subsequently divided into non-overlapping 10-second epochs to facilitate affective state analysis.
\subsubsection{PhysioNet-MI Dataset}
This is a classical motor imagery dataset comprising recordings from 109 participants, acquired using a 64-channel system at an initial sampling rate of 160 Hz. The dataset encompasses four motor imagery tasks: left fist, right fist, both fists, and both feet. The original signals were segmented into 9,837 trials, each lasting 4 seconds.
\subsubsection{BCIC-IV-2a Dataset}
This is a motor imagery dataset comprising recordings from nine subjects performing four distinct imagery tasks: left hand, right hand, both feet, and tongue. Data were acquired over two separate sessions using a 22-channel Ag/AgCl electrode montage at a sampling rate of 250 Hz. Each session contains 288 trials, evenly distributed across the four classes. For preprocessing, we extract the 2–6 second interval from each trial and apply a band-pass filter with cutoff frequencies of 0.3 Hz and 45 Hz to attenuate noise and non-neural artifacts. This yields a total of 5,088 four-second EEG segments for analysis. Following a  subject-independent split, we utilize subjects 1–5 for training, 6–7 for validation, and 8–9 for testing. Notably, although both BCIC-IV-2a and PhysioNet-MI involve four-class MI tasks, \textbf{they differ in both label space and data acquisition paradigms}. 
\subsubsection{ISRUC Dataset}
This is a publicly available sleep dataset that includes three distinct subgroups. We specifically utilize Subgroup I of the ISRUC-Sleep dataset, which consists of 100 full-night polysomnography (PSG) recordings collected from adult participants. Electroencephalographic (EEG) signals were recorded from six standard channels(F3–A2, C3–A2, O1–A2, F4–A1, C4–A1, and O2–A1) at a sampling rate of 200 Hz. The recordings were partitioned into non-overlapping 30-second epochs, each of which was manually scored by certified sleep technicians into one of five sleep stages: Wake, N1, N2, N3, and REM, following the guidelines established by the American Academy of Sleep Medicine (AASM)\cite{berry2017aasm}.
\subsubsection{BCIC2020-3 Dataset}
This is a dataset designed for imagined speech classification, includes EEG recordings from 15 participants who silently articulated one of five phrases—“hello” “help me” “stop” “thank you” or “yes”—without vocalization or movement. Data were acquired using a 64-channel system at a sampling rate of 256 Hz.
\subsubsection{Mumtaz2016 Dataset}
This is a dataset comprising recordings from 34 patients diagnosed with major depressive disorder (MDD) and 30 healthy controls (NCs). EEG signals were acquired from 19 electrodes positioned according to the international 10–20 system at a sampling rate of 256 Hz. The original protocol includes three recording sessions: eyes-open, eyes-closed, and a task-based session. In this work, only the eyes-open and eyes-closed resting-state sessions are utilized.

For all EEG-based downstream tasks, the signals were resampled to 200 Hz. Except for the BCIC-IV-2a dataset, all other datasets were processed and partitioned following the same protocol as CbraMod \cite{wang2024cbramod} to ensure consistent and comparable experimental conditions.

\subsection{Baseline Systems}
To validate the effectiveness and transferability of our method, we conduct comprehensive evaluations on three state-of-the-art EEG foundation models, encompassing both Transformer-based and Mamba-based backbone architectures.

\textbf{CBraMod} \cite{wang2024cbramod} is an EEG foundation model that employs a specialized criss-cross Transformer architecture to capture the distinct spatial and temporal dependencies inherent in EEG signals. It's backbone incorporates two parallel attention pathways: one dedicated to modeling inter-patch spatial dependencies and the other to capturing temporal dynamics. Furthermore, CBraMod introduces an asymmetric positional encoding scheme to effectively encode channel-wise positional information across heterogeneous EEG signals. It is pretrained on a large-scale dataset using a patch-masked reconstruction strategy and achieves state-of-the-art performance across multiple downstream tasks after task-specific fine-tuning.

\textbf{EEGMamba} \cite{wang2025eegmamba} is an EEG foundation model that leverages a bidirectional Mamba-based architecture to effectively capture the complex spatiotemporal dependencies in EEG signals. Its backbone consists of stacked forward and backward Mamba blocks built upon the selective state space model, enabling efficient linear-time sequence modeling while capturing both forward and backward contextual dynamics. EEGMamba employs a dual-domain patch encoder that fuses time-frequency domain features and incorporates conditional positional encoding for adaptive positional awareness. The model is pretrained on a large-scale, heterogeneous EEG corpus using a patch-based masked reconstruction objective. 

\textbf{CSBrain} \cite{zhou2025csbrain} is a cross-scale EEG foundation model that leverages a hierarchical architecture to capture diverse spatiotemporal neural patterns. Its backbone alternately stacks Cross-scale Spatiotemporal Tokenization (CST) to aggregate multi-scale features into scale-aware tokens, and Structured Sparse Attention (SSA) to model long-range cross-window and cross-region dependencies. By replacing dense attention with structured sparsity, CSBrain progressively integrates these dependencies to eliminate spurious correlations and enhance generalization across heterogeneous decoding tasks.

\textbf{DeeperBrain} \cite{wang2026deeper} is a neurophysiologically informed foundation model that embeds biophysical principles into both its architecture and pretraining objectives. It incorporates specific inductive biases to model volume conduction and adaptation dynamics through specialized spatiotemporal encodings. Furthermore, DeeperBrain employs a dual-objective strategy combining Masked EEG Reconstruction (MER) and Neurodynamics Statistics Prediction (NSP), which aligns learned representations with intrinsic neural mechanisms—such as functional connectivity and dynamic complexity—to ensure robust generalization across diverse decoding tasks.

We integrate these baseline foundation models into our pipeline using the official implementations and pre-trained weights provided by their authors. 
\subsection{Procedure}
In our setting, the pretrained foundation model $\mathcal{M}_0$ sequentially adapts to all tasks in multiple fixed orders to evaluate its robustness to the input sequence. 
Each task’s data is split into training, validation, and test sets. Specifically, for the $i$-th task, the model $\mathcal{M}_{i-1}$ is trained on the training set under identical conditions, and the best checkpoint is selected based on validation performance. The corresponding test performance is then reported. This selected model becomes $\mathcal{M}_{i}$ and serves as the starting point for subsequent continual learning. Following adaptation to a new task, we conduct a comprehensive evaluation focusing on two critical properties: the learning plasticity, reflected by the performance on the current task, and the memory stability, measured by the performance on all previously encountered tasks.

\subsection{Evaluation Metrics}
Consistent with existing studies, we adopt the following metrics for evaluation:

\noindent \textbf{Balanced Accuracy (B-ACC)} measures the average of recall obtained on each class, providing an equitable assessment under class imbalance.
\noindent \textbf{Weighted F1 Score (WF1)} computes the F1 score for each class and averages them weighted by support, accounting for label distribution skew.
\noindent \textbf{Cohen’s Kappa ($\kappa$)} quantifies inter-rater agreement beyond chance, reflecting the reliability of classification against random labeling.
\noindent \textbf{AUROC} evaluates the trade-off between true positive and false positive rates across all classification thresholds.
\noindent \textbf{AUC-PR} summarizes precision–recall performance, emphasizing behavior under positive-class rarity.

For multi-class classification, we adopt BACC, WF1, and $\kappa$ as evaluation metrics. For binary classification, we use B-ACC, AUC-PR and AUROC as evaluation metrics. Here, the learning plasticity can be measured by the evaluation metric of model $\mathcal{M}_i$ on the $i$-th task. For memory stability, we quantify the memory retention of the $i$-th task by the \textbf{Relative Forgetting Rate (RFR)}, defined as:
\begin{equation}
\text{RFR} = \frac{\text{A}_{i,i} - \text{A}_{i,\mathcal{N}}}{\text{A}_{i,i}}
\end{equation}
where $\text{A}_{i,i}$ and $\text{A}_{i,\mathcal{N}}$ denote the BACC of model $\mathcal{M}_i$ on the $i$-th task immediately after learning it and that of model $\mathcal{M}_\mathcal{N}$ on the $i$-th task after learning all $\mathcal{N}$ tasks, respectively. A  lower value indicates better retention, with negative values implying performance improvement due to positive transfer—a benefit of cross-task learning.

\begin{table}[!htb]
	\centering
	\caption{Hyper-parameters of EvoBrain and detailed training configurations.}
	\resizebox{1.0\columnwidth}{!}{
		\begin{tabular}{llc}
			\toprule[1pt]
			Category & Hyper-parameter & Value \\
			\midrule
			\multirow{10}{*}{Training}
			& Epochs & 20 \\
			& Batch size & 32 \\
			& Backbone learning rate & $2\times10^{-6}$ \\
			& NSN parameter learning rate & $1\times10^{-3}$ \\
			& Classifier learning rate & $5\times10^{-4}$ \\
			& Weight decay & $5\times10^{-2}$ \\
			& Dropout & 0.1 \\
			& Gradient clipping norm & 0.5 \\
			& Optimizer & AdamW, $\beta=(0.5,0.99)$ \\
			& Label smoothing & 0.1 \\
			\midrule
			\multirow{5}{*}{NSN}
			& Moment-gate initialization $(\eta_{\mu}, \eta_{\sigma})$ & 5.0 \\
			& Band-gate initialization $(a_b)$ & 0.0 \\
			& Spectral residual strength $(\alpha_m)$ & 1.0 \\
			& Band-mask transition width $(\tau)$ & 1 Hz \\
			& Frequency bands $(\mathcal{B})$ & $\delta,\theta,\alpha,\sigma,\beta,\gamma$ \\
			\midrule
			\multirow{3}{*}{RAD}
			& Classification-loss weight $(\lambda_{\mathrm{ce}})$ & 0.1 \\
			& RGD/SAA trade-off weight $(\lambda_r)$ & 0.5 \\
			& Covariance alignment weight $(\rho)$ & 0.3 \\
			\bottomrule[1pt]
	\end{tabular}}
	\label{tab:hyperparameters}
\end{table}
\subsection{Implementation Details}
Our model is trained on a single machine equipped with an Intel Core i9 10900K CPU and eight NVIDIA RTX 3080 GPUs. The architecture of task-specific classifiers are identical to that of CBraMod. Implementation details are provided in Table \ref{tab:hyperparameters}. 

\section{Main Results}
\subsection{Overview Performance}
\begin{figure*}[!bt]
	\centering
	\includegraphics[width=1.0\textwidth]{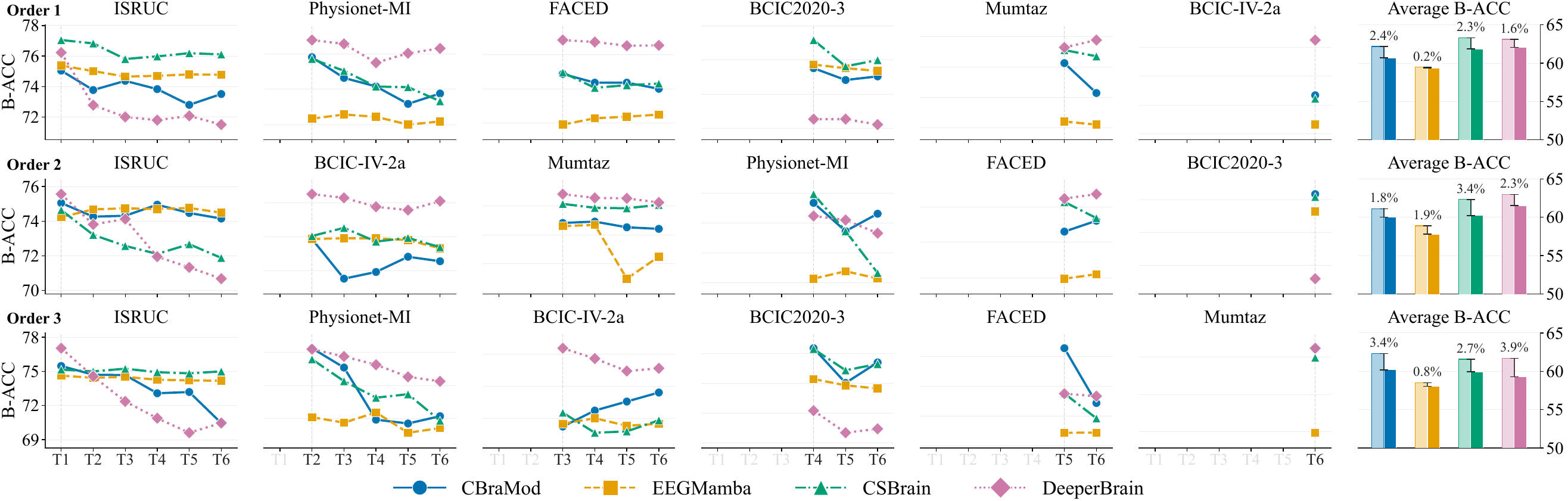}
	\caption{Performance evolution of cross-task continual learning across three distinct task orders. The first six columns depict the B-ACC trajectories for individual datasets as new tasks are sequentially introduced (T1–T6). The rightmost column summarizes the overall stability: the hatched (left) and solid (right) bars indicate the mean initial and final B-ACC across all tasks, respectively, with the annotated gap representing the Relative Forgetting Rate (RFR).}
	\label{fig:main_result}
\end{figure*}

We evaluate EvoBrain across four EEG foundation models in a heterogeneous cross-task continual learning setting, with three task arrival orders reported in Tab. \ref{tab:main_result}. We denote the performance of the $i$-th task immediately after adaptation as $\mathcal{M}_i$, and its performance after completing the full task sequence as $\mathcal{M}_\mathcal{N}$. Fig. \ref{fig:main_result} further tracks the B-ACC trajectory of each task and summarizes the average initial-to-final performance gap using the Relative Forgetting Rate (RFR). Based on these results, we analyze EvoBrain from three perspectives: cross-task learning ability, cross-architecture compatibility, and cross-order robustness.

\subsubsection{Cross-Task Learning Ability}
The results show that EvoBrain can adapt to heterogeneous EEG tasks while maintaining stable performance on previously learned tasks. This behavior is mainly supported by Neuro-Spectral Task Normalization (NSN), which addresses both temporal statistical drift and task-dependent spectral response drift. Specifically, Prototype-guided Temporal Normalization (PTN) stabilizes cross-task representation statistics using historical task prototypes, while Spectral Residual Modulation (SRM) recalibrates frequency-band responses according to task-conditioned spectral gates. These two designs allow the model to fit newly arriving tasks with distinct signal distributions and label spaces. Meanwhile, Response-Affinity Distillation (RAD) preserves historical knowledge by combining Response-Geometry Distillation (RGD), which maintains old-task response relations, and Spectral-Affinity Alignment (SAA), which aligns transferable response statistics only when tasks share compatible spectral signatures. As a result, EvoBrain keeps the average RFR within a low range across all tested settings. In several cases, the final performance even exceeds the immediate post-adaptation performance, such as EEGMamba on FACED in Order 1 improving from 0.439 to 0.450, and CBraMod on BCIC-IV-2a in Order 3 improving from 0.470 to 0.512. These observations suggest that EvoBrain not only mitigates forgetting, but can also exploit beneficial transfer among spectrally related EEG tasks.

\subsubsection{Cross-Architecture Compatibility}
EvoBrain exhibits consistent compatibility across CBraMod, EEGMamba, CSBrain, and DeeperBrain, despite their differences in network structures, pre-training corpora, and representation mechanisms. As shown in Tab. \ref{tab:main_result}, all four backbones retain narrow gaps between their immediate and final average B-ACC values across the three task orders. The average RFR remains below 4.0\% in every backbone-order combination, with the largest degradation being 3.9\% for DeeperBrain under Order 3. This indicates that EvoBrain does not rely on a specific backbone architecture to balance plasticity and stability. Instead, NSN provides a task-aware normalization interface for adapting heterogeneous EEG representations, while RAD regularizes the continual update process through response-level structure preservation and affinity-guided transfer. Therefore, the framework can be integrated into diverse EEG foundation models without requiring structural modifications.

\subsubsection{Cross-Order Robustness}
We further validate the robustness of the framework against sequential variations by examining performance across three distinct sequential permutations (Orders 1–3). Across different orders, the final Average B-ACC remains stable across task orders for all backbones: 
CBraMod obtains $0.603 \pm 0.003$, 
EEGMamba obtains $0.584 \pm 0.007$, 
CSBrain obtains $0.607 \pm 0.009$, 
and DeeperBrain obtains $0.610 \pm 0.012$. Although different task permutations introduce different degrees of difficulty, the corresponding RFR values remain low, ranging from 0.1\% to 3.9\% across all settings. This controlled order sensitivity is enabled by the complementary roles of NSN and RAD. NSN reduces the dependence of optimization on the statistical properties of early-arriving tasks, whereas RAD constrains the response geometry of replayed historical samples and promotes transfer only between tasks with compatible spectral affinity. Such robustness is important for practical BCI scenarios, where calibration tasks may appear in unpredictable sequences and continual adaptation must remain stable without retraining from scratch.
\begin{table*}[!tb]
	\centering
	\caption{Overview performance of cross-task continual learning under different task input orders across EEG foundation models. WF1 and $\kappa$ are used for multi-class tasks; AUROC and AUC-PR for binary tasks. $\mathcal{M}_i$ denotes the performance immediately after adapting to the $i$-th task, whereas $\mathcal{M}_\mathcal{N}$ represents the performance after adapting to the final task. }
	\resizebox{1.0\textwidth}{!}{%
		\begin{tabular}{llccccccccccccccccccccc}
			\toprule
			\multicolumn{2}{c}{\multirow{3}{*}{\textbf{Order 1}}} & \multicolumn{3}{c}{ISRUC, 5-classes} & \multicolumn{3}{c}{Physionet-MI, 4-classes} & \multicolumn{3}{c}{FACED, 9-classes} & \multicolumn{3}{c}{BCIC-2020-3, 5-classes} & \multicolumn{3}{c}{Mumtaz, 2-classes} & \multicolumn{3}{c}{BCIC-IV-2a, 4-classes} & \multicolumn{3}{c}{\multirow{2}{*}{Average B-ACC}} \\
			&  & \multicolumn{3}{c}{[Sleep Staging]} & \multicolumn{3}{c}{[Motor Imagery]} & \multicolumn{3}{c}{[Emotion Recognition]} & \multicolumn{3}{c}{[Imagined Speech]} & \multicolumn{3}{c}{[Mental Dis. Diag.]} & \multicolumn{3}{c}{[Motor Imagery]} & \multicolumn{3}{c}{} \\
			\cmidrule(lr){3-5} \cmidrule(lr){6-8} \cmidrule(lr){9-11} \cmidrule(lr){12-14} \cmidrule(lr){15-17} \cmidrule(lr){18-20} \cmidrule(lr){21-23}
			\multicolumn{2}{c}{} & $\mathcal{M}_i$ & $\mathcal{M}_\mathcal{N}$ & RFR & $\mathcal{M}_i$ & $\mathcal{M}_\mathcal{N}$ & RFR & $\mathcal{M}_i$ & $\mathcal{M}_\mathcal{N}$ & RFR & $\mathcal{M}_i$ & $\mathcal{M}_\mathcal{N}$ & RFR & $\mathcal{M}_i$ & $\mathcal{M}_\mathcal{N}$ & RFR & $\mathcal{M}_i$ & $\mathcal{M}_\mathcal{N}$ & RFR & $\mathcal{M}_i$ & $\mathcal{M}_\mathcal{N}$ & RFR \\
			\midrule
			\multirow{3}{*}{CBraMod} 
			& B-ACC           & 0.751 & 0.735 & 2.0\% & 0.573 & 0.547 & 4.5\% & 0.499 & 0.480 & 3.5\% & 0.488 & 0.476 & 2.5\% & 0.902 & 0.883 & 2.1\% & 0.520 & 0.520 & - & \multirow{3}{*}{0.622} & \multirow{3}{*}{0.607} & \multirow{3}{*}{2.4\%} \\
			& $\kappa$/AUC-PR & 0.697 & 0.672 & 3.7\% & 0.430 & 0.396 & 7.9\% & 0.431 & 0.411 & 4.5\% & 0.360 & 0.345 & 4.1\% & 0.960 & 0.922 & 3.9\% & 0.360 & 0.360 & - & & & \\
			& WF1/AUROC       & 0.763 & 0.733 & 3.9\% & 0.569 & 0.546 & 4.0\% & 0.492 & 0.477 & 3.0\% & 0.488 & 0.476 & 2.4\% & 0.969 & 0.951 & 1.8\% & 0.514 & 0.514 & - & & & \\
			\midrule
			\multirow{3}{*}{EEGMamba} 
			& B-ACC           & 0.754 & 0.748 & 0.8\% & 0.529 & 0.527 & 0.4\% & 0.439 & 0.450 & -2.6\% & 0.493 & 0.484 & 1.9\% & 0.864 & 0.862 & 0.2\% & 0.490 & 0.490 & - & \multirow{3}{*}{0.595} & \multirow{3}{*}{0.594} & \multirow{3}{*}{0.1\%} \\
			& $\kappa$/AUC-PR & 0.698 & 0.685 & 1.8\% & 0.372 & 0.369 & 0.8\% & 0.368 & 0.378 & -2.7\% & 0.367 & 0.355 & 3.2\% & 0.925 & 0.921 & 0.4\% & 0.319 & 0.319 & - & & & \\
			& WF1/AUROC       & 0.765 & 0.757 & 1.0\% & 0.524 & 0.515 & 1.6\% & 0.439 & 0.445 & -1.4\% & 0.493 & 0.483 & 2.1\% & 0.921 & 0.915 & 0.6\% & 0.473 & 0.473 & - & & & \\
			\midrule
			\multirow{3}{*}{CSBrain} 
			& B-ACC           & 0.771 & 0.761 & 1.2\% & 0.571 & 0.541 & 5.2\% & 0.499 & 0.486 & 2.6\% & 0.529 & 0.500 & 5.5\% & 0.911 & 0.907 & 0.4\% & 0.516 & 0.516 & - & \multirow{3}{*}{0.633} & \multirow{3}{*}{0.619} & \multirow{3}{*}{2.3\%} \\
			& $\kappa$/AUC-PR & 0.709 & 0.705 & 0.6\% & 0.428 & 0.389 & 9.3\% & 0.436 & 0.418 & 4.2\% & 0.412 & 0.375 & 8.9\% & 0.980 & 0.979 & 0.1\% & 0.355 & 0.355 & - & & & \\
			& WF1/AUROC       & 0.775 & 0.773 & 0.2\% & 0.573 & 0.540 & 5.9\% & 0.508 & 0.482 & 5.1\% & 0.529 & 0.498 & 5.9\% & 0.978 & 0.977 & 0.1\% & 0.506 & 0.506 & - & & & \\
			\midrule
			\multirow{3}{*}{DeeperBrain} 
			& B-ACC           & 0.762 & 0.715 & 6.2\% & 0.585 & 0.579 & 1.0\% & 0.537 & 0.530 & 1.2\% & 0.413 & 0.405 & 1.9\% & 0.913 & 0.918 & -0.5\% & 0.577 & 0.577 & - & \multirow{3}{*}{0.631} & \multirow{3}{*}{0.621} & \multirow{3}{*}{1.6\%} \\
			& $\kappa$/AUC-PR & 0.731 & 0.666 & 9.0\% & 0.446 & 0.438 & 1.8\% & 0.472 & 0.462 & 2.1\% & 0.267 & 0.257 & 3.8\% & 0.978 & 0.980 & -0.2\% & 0.436 & 0.436 & - & & & \\
			& WF1/AUROC       & 0.784 & 0.722 & 7.9\% & 0.585 & 0.575 & 1.7\% & 0.530 & 0.512 & 3.3\% & 0.414 & 0.406 & 1.9\% & 0.976 & 0.978 & -0.2\% & 0.568 & 0.568 & - & & & \\
			\bottomrule
		\end{tabular}%
	}
	
	\vspace{0.2cm}
	\resizebox{1.0\textwidth}{!}{%
		\begin{tabular}{llccccccccccccccccccccc}
			\toprule
			\multicolumn{2}{c}{\multirow{2}{*}{\textbf{Order 2}}} & \multicolumn{3}{c}{ISRUC, 5-classes} & \multicolumn{3}{c}{BCIC-IV-2a, 4-classes} & \multicolumn{3}{c}{Mumtaz, 2-classes} & \multicolumn{3}{c}{Physionet-MI, 4-classes} & \multicolumn{3}{c}{FACED, 9-classes} & \multicolumn{3}{c}{BCIC-2020-3, 5-classes} & \multicolumn{3}{c}{\multirow{2}{*}{Average B-ACC}} \\
			&  & \multicolumn{3}{c}{[Sleep Staging]} & \multicolumn{3}{c}{[Motor Imagery]} & \multicolumn{3}{c}{[Mental Dis. Diag.]} & \multicolumn{3}{c}{[Motor Imagery]} & \multicolumn{3}{c}{[Emotion Recognition]} & \multicolumn{3}{c}{[Imagined Speech]} & \multicolumn{3}{c}{} \\
			\cmidrule(lr){3-5} \cmidrule(lr){6-8} \cmidrule(lr){9-11} \cmidrule(lr){12-14} \cmidrule(lr){15-17} \cmidrule(lr){18-20} \cmidrule(lr){21-23}
			\multicolumn{2}{c}{} & $\mathcal{M}_i$ & $\mathcal{M}_\mathcal{N}$ & RFR & $\mathcal{M}_i$ & $\mathcal{M}_\mathcal{N}$ & RFR & $\mathcal{M}_i$ & $\mathcal{M}_\mathcal{N}$ & RFR & $\mathcal{M}_i$ & $\mathcal{M}_\mathcal{N}$ & RFR & $\mathcal{M}_i$ & $\mathcal{M}_\mathcal{N}$ & RFR & $\mathcal{M}_i$ & $\mathcal{M}_\mathcal{N}$ & RFR & $\mathcal{M}_i$ & $\mathcal{M}_\mathcal{N}$ & RFR \\
			\midrule
			\multirow{3}{*}{CBraMod} 
			& B-ACC           & 0.751 & 0.741 & 1.2\% & 0.475 & 0.423 & 10.9\% & 0.871 & 0.860 & 1.2\% & 0.594 & 0.586 & 1.2\% & 0.492 & 0.504 & -2.5\% & 0.485 & 0.485 & - & \multirow{3}{*}{0.611} & \multirow{3}{*}{0.600} & \multirow{3}{*}{1.8\%} \\
			& $\kappa$/AUC-PR & 0.697 & 0.685 & 1.8\% & 0.300 & 0.230 & 23.2\% & 0.966 & 0.889 & 7.9\% & 0.458 & 0.448 & 2.1\% & 0.427 & 0.441 & -3.4\% & 0.357 & 0.357 & - & & & \\
			& WF1/AUROC       & 0.763 & 0.749 & 1.9\% & 0.444 & 0.362 & 18.4\% & 0.964 & 0.920 & 4.5\% & 0.596 & 0.587 & 1.6\% & 0.490 & 0.507 & -3.5\% & 0.485 & 0.485 & - & & & \\
			\midrule
			\multirow{3}{*}{EEGMamba} 
			& B-ACC           & 0.742 & 0.745 & -0.4\% & 0.475 & 0.453 & 4.6\% & 0.865 & 0.813 & 6.1\% & 0.543 & 0.543 & 0.0\% & 0.439 & 0.444 & -1.1\% & 0.469 & 0.469 & - & \multirow{3}{*}{0.589} & \multirow{3}{*}{0.578} & \multirow{3}{*}{1.9\%} \\
			& $\kappa$/AUC-PR & 0.681 & 0.689 & -1.1\% & 0.300 & 0.271 & 9.7\% & 0.934 & 0.916 & 2.0\% & 0.390 & 0.390 & 0.0\% & 0.370 & 0.374 & -1.3\% & 0.337 & 0.337 & - & & & \\
			& WF1/AUROC       & 0.747 & 0.754 & -1.0\% & 0.476 & 0.446 & 6.2\% & 0.918 & 0.903 & 1.6\% & 0.546 & 0.540 & 1.1\% & 0.441 & 0.444 & -0.7\% & 0.468 & 0.468 & - & & & \\
			\midrule
			\multirow{3}{*}{CSBrain} 
			& B-ACC           & 0.746 & 0.719 & 3.7\% & 0.482 & 0.456 & 5.4\% & 0.903 & 0.902 & 0.2\% & 0.600 & 0.546 & 8.9\% & 0.526 & 0.507 & 3.5\% & 0.483 & 0.483 & - & \multirow{3}{*}{0.623} & \multirow{3}{*}{0.602} & \multirow{3}{*}{3.4\%} \\
			& $\kappa$/AUC-PR & 0.685 & 0.644 & 5.9\% & 0.309 & 0.274 & 11.2\% & 0.982 & 0.979 & 3.6\% & 0.466 & 0.395 & 15.2\% & 0.466 & 0.444 & 4.8\% & 0.353 & 0.353 & - & & & \\
			& WF1/AUROC       & 0.747 & 0.719 & 3.8\% & 0.466 & 0.448 & 3.9\% & 0.982 & 0.979 & 0.3\% & 0.604 & 0.552 & 8.6\% & 0.539 & 0.516 & 4.3\% & 0.483 & 0.483 & - & & & \\
			\midrule
			\multirow{3}{*}{DeeperBrain} 
			& B-ACC           & 0.756 & 0.707 & 6.4\% & 0.580 & 0.563 & 2.8\% & 0.920 & 0.906 & 1.6\% & 0.585 & 0.573 & 2.0\% & 0.529 & 0.534 & -1.0\% & 0.408 & 0.408 & - & \multirow{3}{*}{0.630} & \multirow{3}{*}{0.615} & \multirow{3}{*}{2.3\%} \\
			& $\kappa$/AUC-PR & 0.703 & 0.638 & 9.2\% & 0.440 & 0.418 & 5.0\% & 0.977 & 0.976 & 0.1\% & 0.446 & 0.431 & 3.5\% & 0.469 & 0.473 & -0.1\% & 0.260 & 0.260 & - & & & \\
			& WF1/AUROC       & 0.770 & 0.719 & 6.7\% & 0.567 & 0.545 & 4.0\% & 0.976 & 0.977 & -0.1\% & 0.587 & 0.573 & 2.3\% & 0.532 & 0.533 & -0.1\% & 0.406 & 0.406 & - & & & \\
			\bottomrule
		\end{tabular}%
	}
	
	\vspace{0.2cm} 
	\resizebox{1.0\textwidth}{!}{%
		\begin{tabular}{llccccccccccccccccccccc}
			\toprule
			\multicolumn{2}{c}{\multirow{2}{*}{\textbf{Order 3}}} & \multicolumn{3}{c}{ISRUC, 5-classes} & \multicolumn{3}{c}{Physionet-MI, 4-classes} & \multicolumn{3}{c}{BCIC-IV-2a, 4-classes} & \multicolumn{3}{c}{BCIC-2020-3, 5-classes} & \multicolumn{3}{c}{FACED, 9-classes} & \multicolumn{3}{c}{Mumtaz, 2-classes} & \multicolumn{3}{c}{\multirow{2}{*}{Average B-ACC}} \\
			&  & \multicolumn{3}{c}{[Sleep Staging]} & \multicolumn{3}{c}{[Motor Imagery]} & \multicolumn{3}{c}{[Motor Imagery]} & \multicolumn{3}{c}{[Imagined Speech]} & \multicolumn{3}{c}{[Emotion Recognition]} & \multicolumn{3}{c}{[Mental Dis. Diag.]} & \multicolumn{3}{c}{} \\
			\cmidrule(lr){3-5} \cmidrule(lr){6-8} \cmidrule(lr){9-11} \cmidrule(lr){12-14} \cmidrule(lr){15-17} \cmidrule(lr){18-20} \cmidrule(lr){21-23}
			\multicolumn{2}{c}{} & $\mathcal{M}_i$ & $\mathcal{M}_\mathcal{N}$ & RFR & $\mathcal{M}_i$ & $\mathcal{M}_\mathcal{N}$ & RFR & $\mathcal{M}_i$ & $\mathcal{M}_\mathcal{N}$ & RFR & $\mathcal{M}_i$ & $\mathcal{M}_\mathcal{N}$ & RFR & $\mathcal{M}_i$ & $\mathcal{M}_\mathcal{N}$ & RFR & $\mathcal{M}_i$ & $\mathcal{M}_\mathcal{N}$ & RFR & $\mathcal{M}_i$ & $\mathcal{M}_\mathcal{N}$ & RFR \\
			\midrule
			\multirow{3}{*}{CBraMod} 
			& B-ACC           & 0.755 & 0.705 & 6.6\% & 0.582 & 0.541 & 7.0\% & 0.470 & 0.512 & -9.1\% & 0.509 & 0.485 & 4.7\% & 0.518 & 0.463 & 10.6\% & 0.904 & 0.904 & - & \multirow{3}{*}{0.623} & \multirow{3}{*}{0.602} & \multirow{3}{*}{3.4\%} \\
			& $\kappa$/AUC-PR & 0.683 & 0.625 & 8.4\% & 0.443 & 0.388 & 12.3\% & 0.293 & 0.350 & -19.4\% & 0.387 & 0.357 & 7.8\% & 0.454 & 0.392 & 13.7\% & 0.969 & 0.969 & - & & & \\
			& WF1/AUROC       & 0.751 & 0.697 & 7.2\% & 0.581 & 0.536 & 7.6\% & 0.455 & 0.496 & -9.2\% & 0.509 & 0.484 & 4.9\% & 0.522 & 0.470 & 9.9\% & 0.965 & 0.965 & - & & & \\
			\midrule
			\multirow{3}{*}{EEGMamba} 
			& B-ACC           & 0.746 & 0.742 & 0.6\% & 0.541 & 0.534 & 1.2\% & 0.473 & 0.473 & 0.0\% & 0.457 & 0.441 & 3.5\% & 0.433 & 0.433 & 0.0\% & 0.859 & 0.859 & - & \multirow{3}{*}{0.585} & \multirow{3}{*}{0.581} & \multirow{3}{*}{0.8\%} \\
			& $\kappa$/AUC-PR & 0.679 & 0.674 & 0.7\% & 0.388 & 0.379 & 2.3\% & 0.297 & 0.297 & 0.0\% & 0.322 & 0.302 & 6.2\% & 0.363 & 0.364 & -0.1\% & 0.936 & 0.936 & - & & & \\
			& WF1/AUROC       & 0.747 & 0.745 & 0.1\% & 0.542 & 0.526 & 2.9\% & 0.463 & 0.462 & 0.1\% & 0.457 & 0.439 & 4.0\% & 0.434 & 0.436 & -0.4\% & 0.919 & 0.919 & - & & & \\
			\midrule
			\multirow{3}{*}{CSBrain} 
			& B-ACC           & 0.752 & 0.750 & 0.2\% & 0.576 & 0.539 & 6.4\% & 0.487 & 0.477 & 2.0\% & 0.508 & 0.483 & 4.9\% & 0.473 & 0.448 & 5.3\% & 0.900 & 0.900 & - & \multirow{3}{*}{0.616} & \multirow{3}{*}{0.599} & \multirow{3}{*}{2.7\%} \\
			& $\kappa$/AUC-PR & 0.680 & 0.679 & 0.2\% & 0.434 & 0.385 & 11.4\% & 0.316 & 0.303 & 4.0\% & 0.385 & 0.353 & 8.2\% & 0.404 & 0.376 & 6.9\% & 0.972 & 0.972 & - & & & \\
			& WF1/AUROC       & 0.748 & 0.751 & -0.4\% & 0.578 & 0.537 & 7.1\% & 0.485 & 0.475 & 2.1\% & 0.506 & 0.482 & 4.8\% & 0.472 & 0.448 & 5.0\% & 0.966 & 0.966 & - & & & \\
			\midrule
			\multirow{3}{*}{DeeperBrain} 
			& B-ACC           & 0.770 & 0.704 & 8.5\% & 0.582 & 0.562 & 3.3\% & 0.568 & 0.543 & 4.4\% & 0.404 & 0.373 & 7.6\% & 0.473 & 0.470 & 0.5\% & 0.905 & 0.905 & - & \multirow{3}{*}{0.617} & \multirow{3}{*}{0.593} & \multirow{3}{*}{3.9\%} \\
			& $\kappa$/AUC-PR & 0.702 & 0.608 & 13.3\% & 0.442 & 0.417 & 5.8\% & 0.424 & 0.390 & 7.9\% & 0.255 & 0.217 & 15.0\% & 0.401 & 0.398 & 0.8\% & 0.978 & 0.978 & - & & & \\
			& WF1/AUROC       & 0.768 & 0.680 & 11.4\% & 0.583 & 0.563 & 3.5\% & 0.559 & 0.527 & 5.6\% & 0.404 & 0.373 & 7.8\% & 0.474 & 0.473 & 0.2\% & 0.975 & 0.975 & - & & & \\
			\bottomrule
		\end{tabular}%
	}
	\label{tab:main_result}
\end{table*}

\subsection{Comparison with Other Methods}
\begin{figure*}[!tb]
	\centering
	\includegraphics[width=1.0\textwidth]{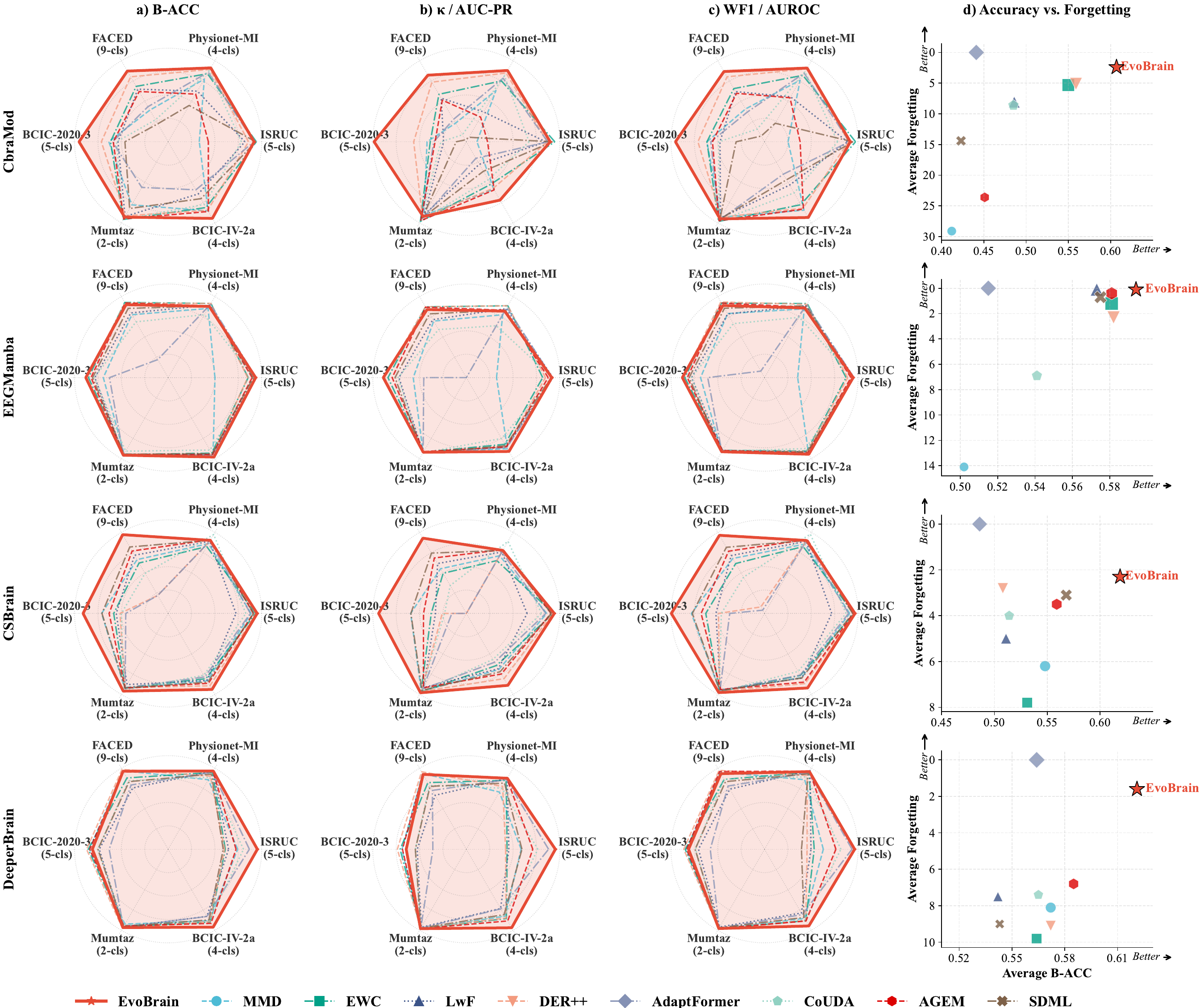}
	\caption{Comparative evaluation of the proposed EvoBrain framework against other methods across four EEG foundation models. The rows correspond to four distinct backbones: CBraMod, EEGMamba, CSBrain, and DeeperBrain. (a)–(c) Radar charts visualize the final retained performance across six heterogeneous tasks in terms of B-ACC, $\kappa$/AUC-PR, and WF1/AUROC, respectively. (d) Accuracy vs. Forgetting scatter plots illustrate the stability-plasticity trade-off. The x-axis represents the Average B-ACC (higher is better), while the y-axis denotes the Average Forgetting (lower is better, plotted inversely). }
	\label{fig:compared_result}
\end{figure*}

\begin{table*}[t]
	\centering
	\caption{Ablation analysis of EvoBrain integrated into diverse EEG foundation model architectures.}
	\label{tab:ablation_study}
	
	\begingroup
	\setlength{\tabcolsep}{3.2pt}
	\renewcommand{\arraystretch}{1.08}
	
	\begin{adjustbox}{max width=\textwidth}
		\begin{tabular}{
				ll
				*{12}{c}
				@{\hspace{1.5pt}}c@{\hspace{1.5pt}}c@{\hspace{1.5pt}}c
				*{6}{c}
			}
			\toprule
			&  & \multicolumn{3}{c}{ISRUC, 5-classes} 
			& \multicolumn{3}{c}{Physionet-MI, 4-classes} 
			& \multicolumn{3}{c}{FACED, 9-classes} 
			& \multicolumn{3}{c}{BCIC-2020-3, 5-classes} 
			& \multicolumn{3}{c}{Mumtaz, 2-classes} 
			& \multicolumn{3}{c}{BCIC-IV-2a, 4-classes} 
			& \multicolumn{3}{c}{Average B-ACC} \\
			&  & B-ACC & $\kappa$ & WF1
			& B-ACC & $\kappa$ & WF1
			& B-ACC & $\kappa$ & WF1
			& B-ACC & $\kappa$ & WF1
			& B-ACC & AUC-PR & AUROC
			& B-ACC & $\kappa$ & WF1
			& $\mathcal{M}_i$ & $\mathcal{M}_\mathcal{N}$ & RFR \\
			\cmidrule(lr){3-5}
			\cmidrule(lr){6-8}
			\cmidrule(lr){9-11}
			\cmidrule(lr){12-14}
			\cmidrule(lr){15-17}
			\cmidrule(lr){18-20}
			\cmidrule(lr){21-23}
			
			\rowcolor{gray!12}
			& EvoBrain
			& 0.735 & 0.672 & 0.733
			& 0.547 & 0.396 & 0.546
			& 0.480 & 0.411 & 0.477
			& 0.476 & 0.345 & 0.476
			& 0.883 & 0.922 & 0.951
			& 0.520 & 0.360 & 0.514
			& 0.622 & \textbf{0.607} & \textbf{2.4\%} \\
			
			\rowcolor{gray!5}
			& w/o NSN
			& 0.743 & 0.697 & 0.760
			& 0.516 & 0.354 & 0.519
			& 0.393 & 0.309 & 0.379
			& 0.359 & 0.198 & 0.350
			& 0.877 & 0.957 & 0.961
			& 0.455 & 0.273 & 0.429
			& 0.590 & 0.557 & 5.6\% \\
			
			& \subab{w/o PTN}
			& 0.754 & 0.700 & 0.761
			& 0.533 & 0.377 & 0.533
			& 0.408 & 0.329 & 0.399
			& 0.357 & 0.197 & 0.355
			& 0.901 & 0.951 & 0.961
			& 0.457 & 0.277 & 0.429
			& 0.583 & 0.569 & 2.5\% \\
			
			& \subab{w/o SRM}
			& 0.704 & 0.645 & 0.716
			& 0.515 & 0.354 & 0.516
			& 0.440 & 0.363 & 0.433
			& 0.444 & 0.305 & 0.443
			& 0.809 & 0.800 & 0.879
			& 0.522 & 0.362 & 0.512
			& 0.613 & 0.572 & 6.7\% \\
			
			\rowcolor{gray!5}
			& w/o RAD
			& 0.731 & 0.679 & 0.742
			& 0.557 & 0.409 & 0.559
			& 0.463 & 0.395 & 0.470
			& 0.437 & 0.297 & 0.434
			& 0.767 & 0.816 & 0.879
			& 0.509 & 0.345 & 0.500
			& 0.605 & 0.577 & 4.6\% \\
			
			& \subab{w/o RGD}
			& 0.723 & 0.670 & 0.736
			& 0.551 & 0.402 & 0.551
			& 0.455 & 0.385 & 0.456
			& 0.451 & 0.313 & 0.449
			& 0.653 & 0.770 & 0.832
			& 0.504 & 0.339 & 0.499
			& 0.608 & 0.556 & 8.6\% \\
			
			& \subab{w/o SAA}
			& 0.732 & 0.667 & 0.731
			& 0.545 & 0.394 & 0.542
			& 0.456 & 0.382 & 0.445
			& 0.464 & 0.330 & 0.462
			& 0.843 & 0.911 & 0.944
			& 0.515 & 0.353 & 0.501
			& 0.618 & 0.592 & 4.2\% \\
			
			\multirow{-8}{*}{CBraMod}
			& w/o NSN+RAD
			& 0.732 & 0.695 & 0.760
			& 0.464 & 0.285 & 0.465
			& 0.353 & 0.271 & 0.344
			& 0.337 & 0.172 & 0.330
			& 0.879 & 0.933 & 0.944
			& 0.458 & 0.278 & 0.431
			& 0.589 & 0.537 & 8.8\% \\
			\specialrule{0.8pt}{2pt}{1pt}
			
			\rowcolor{gray!12}
			& EvoBrain
			& 0.748 & 0.685 & 0.757
			& 0.527 & 0.369 & 0.515
			& 0.450 & 0.378 & 0.445
			& 0.484 & 0.355 & 0.483
			& 0.862 & 0.921 & 0.915
			& 0.490 & 0.319 & 0.473
			& 0.595 & \textbf{0.594} & \textbf{0.1\%} \\
			
			\rowcolor{gray!5}
			& w/o NSN
			& 0.714 & 0.633 & 0.716
			& 0.538 & 0.384 & 0.524
			& 0.450 & 0.376 & 0.446
			& 0.448 & 0.310 & 0.447
			& 0.857 & 0.919 & 0.901
			& 0.475 & 0.300 & 0.470
			& 0.587 & 0.580 & 1.2\% \\
			
			& \subab{w/o PTN}
			& 0.735 & 0.667 & 0.734
			& 0.527 & 0.369 & 0.518
			& 0.442 & 0.370 & 0.441
			& 0.487 & 0.358 & 0.486
			& 0.858 & 0.921 & 0.911
			& 0.483 & 0.311 & 0.464
			& 0.590 & 0.588 & 0.2\% \\
			
			& \subab{w/o SRM}
			& 0.755 & 0.689 & 0.757
			& 0.526 & 0.368 & 0.517
			& 0.447 & 0.374 & 0.440
			& 0.493 & 0.367 & 0.493
			& 0.859 & 0.929 & 0.917
			& 0.484 & 0.311 & 0.465
			& 0.596 & 0.594 & 0.3\% \\
			
			\rowcolor{gray!5}
			& w/o RAD
			& 0.686 & 0.630 & 0.706
			& 0.521 & 0.362 & 0.523
			& 0.443 & 0.371 & 0.444
			& 0.451 & 0.313 & 0.449
			& 0.861 & 0.919 & 0.910
			& 0.491 & 0.322 & 0.485
			& 0.580 & 0.575 & 0.8\% \\
			
			& \subab{w/o RGD}
			& 0.702 & 0.649 & 0.722
			& 0.528 & 0.371 & 0.520
			& 0.413 & 0.339 & 0.413
			& 0.489 & 0.362 & 0.489
			& 0.838 & 0.931 & 0.919
			& 0.490 & 0.319 & 0.472
			& 0.589 & 0.577 & 2.1\% \\
			
			& \subab{w/o SAA}
			& 0.748 & 0.702 & 0.770
			& 0.535 & 0.524 & 0.380
			& 0.451 & 0.381 & 0.449
			& 0.464 & 0.330 & 0.463
			& 0.853 & 0.925 & 0.914
			& 0.483 & 0.310 & 0.474
			& 0.591 & 0.589 & 0.3\% \\
			
			\multirow{-8}{*}{EEGMamba}
			& w/o NSN+RAD
			& 0.695 & 0.632 & 0.712
			& 0.527 & 0.369 & 0.528
			& 0.455 & 0.384 & 0.456
			& 0.489 & 0.362 & 0.489
			& 0.844 & 0.925 & 0.905
			& 0.484 & 0.311 & 0.471
			& 0.590 & 0.582 & 1.3\% \\
			\specialrule{0.8pt}{2pt}{1pt}
			
			\rowcolor{gray!12}
			& EvoBrain
			& 0.761 & 0.705 & 0.773
			& 0.541 & 0.389 & 0.540
			& 0.486 & 0.418 & 0.482
			& 0.500 & 0.375 & 0.498
			& 0.911 & 0.979 & 0.977
			& 0.516 & 0.355 & 0.506
			& 0.633 & \textbf{0.619} & 2.3\% \\
			
			\rowcolor{gray!5}
			& w/o NSN
			& 0.761 & 0.717 & 0.785
			& 0.557 & 0.409 & 0.557
			& 0.436 & 0.364 & 0.438
			& 0.399 & 0.248 & 0.399
			& 0.905 & 0.977 & 0.977
			& 0.508 & 0.344 & 0.493
			& 0.606 & 0.593 & 2.2\% \\
			
			& \subab{w/o PTN}
			& 0.729 & 0.666 & 0.755
			& 0.535 & 0.380 & 0.536
			& 0.457 & 0.385 & 0.454
			& 0.445 & 0.307 & 0.445
			& 0.905 & 0.968 & 0.963
			& 0.506 & 0.341 & 0.497
			& 0.613 & 0.596 & 2.7\% \\
			
			& \subab{w/o SRM}
			& 0.745 & 0.679 & 0.756
			& 0.549 & 0.398 & 0.546
			& 0.470 & 0.398 & 0.464
			& 0.479 & 0.348 & 0.477
			& 0.898 & 0.967 & 0.963
			& 0.498 & 0.331 & 0.492
			& 0.621 & 0.607 & 2.3\% \\
			
			\rowcolor{gray!5}
			& w/o RAD
			& 0.747 & 0.684 & 0.757
			& 0.512 & 0.349 & 0.514
			& 0.460 & 0.386 & 0.460
			& 0.463 & 0.328 & 0.463
			& 0.891 & 0.971 & 0.964
			& 0.509 & 0.346 & 0.502
			& 0.631 & 0.597 & 5.4\% \\
			
			& \subab{w/o RGD}
			& 0.722 & 0.660 & 0.746
			& 0.522 & 0.362 & 0.518
			& 0.476 & 0.405 & 0.467
			& 0.473 & 0.342 & 0.468
			& 0.908 & 0.968 & 0.960
			& 0.467 & 0.289 & 0.464
			& 0.622 & 0.595 & 4.4\% \\
			
			& \subab{w/o SAA}
			& 0.749 & 0.686 & 0.763
			& 0.531 & 0.374 & 0.529
			& 0.482 & 0.413 & 0.480
			& 0.485 & 0.357 & 0.485
			& 0.891 & 0.963 & 0.959
			& 0.502 & 0.336 & 0.496
			& 0.619 & 0.607 & \textbf{2.0\%} \\
			
			\multirow{-8}{*}{CSBrain}
			& w/o NSN+RAD
			& 0.708 & 0.625 & 0.718
			& 0.553 & 0.404 & 0.551
			& 0.350 & 0.268 & 0.348
			& 0.304 & 0.130 & 0.298
			& 0.886 & 0.969 & 0.969
			& 0.487 & 0.316 & 0.471
			& 0.585 & 0.548 & 6.4\% \\
			\specialrule{0.8pt}{2pt}{1pt}
			
			\rowcolor{gray!12}
			& EvoBrain
			& 0.715 & 0.666 & 0.722
			& 0.579 & 0.438 & 0.575
			& 0.530 & 0.462 & 0.513
			& 0.405 & 0.257 & 0.406
			& 0.918 & 0.980 & 0.978
			& 0.577 & 0.436 & 0.568
			& 0.631 & \textbf{0.621} & \textbf{1.6\%} \\
			
			\rowcolor{gray!5}
			& w/o NSN
			& 0.602 & 0.507 & 0.586
			& 0.543 & 0.391 & 0.545
			& 0.521 & 0.453 & 0.513
			& 0.453 & 0.317 & 0.452
			& 0.872 & 0.977 & 0.977
			& 0.564 & 0.419 & 0.558
			& 0.627 & 0.593 & 5.6\% \\
			
			& \subab{w/o PTN}
			& 0.637 & 0.544 & 0.612
			& 0.544 & 0.392 & 0.544
			& 0.513 & 0.441 & 0.492
			& 0.419 & 0.273 & 0.416
			& 0.902 & 0.976 & 0.975
			& 0.563 & 0.418 & 0.551
			& 0.622 & 0.596 & 4.1\% \\
			
			& \subab{w/o SRM}
			& 0.691 & 0.637 & 0.705
			& 0.557 & 0.409 & 0.556
			& 0.514 & 0.447 & 0.508
			& 0.376 & 0.220 & 0.369
			& 0.908 & 0.978 & 0.977
			& 0.563 & 0.417 & 0.555
			& 0.617 & 0.601 & 2.6\% \\
			
			\rowcolor{gray!5}
			& w/o RAD
			& 0.561 & 0.431 & 0.498
			& 0.531 & 0.374 & 0.524
			& 0.536 & 0.472 & 0.529
			& 0.436 & 0.295 & 0.435
			& 0.912 & 0.982 & 0.982
			& 0.574 & 0.432 & 0.569
			& 0.632 & 0.592 & 6.4\% \\
			
			& \subab{w/o RGD}
			& 0.629 & 0.534 & 0.609
			& 0.563 & 0.417 & 0.564
			& 0.514 & 0.445 & 0.499
			& 0.444 & 0.305 & 0.442
			& 0.894 & 0.975 & 0.975
			& 0.582 & 0.442 & 0.573
			& 0.629 & 0.604 & 3.9\% \\
			
			& \subab{w/o SAA}
			& 0.629 & 0.546 & 0.623
			& 0.558 & 0.410 & 0.556
			& 0.528 & 0.462 & 0.527
			& 0.429 & 0.287 & 0.424
			& 0.901 & 0.977 & 0.976
			& 0.559 & 0.412 & 0.553
			& 0.633 & 0.600 & 5.1\% \\
			
			\multirow{-8}{*}{DeeperBrain}
			& w/o NSN+RAD
			& 0.535 & 0.399 & 0.468
			& 0.532 & 0.376 & 0.532
			& 0.523 & 0.459 & 0.518
			& 0.419 & 0.273 & 0.417
			& 0.897 & 0.979 & 0.977
			& 0.556 & 0.409 & 0.548
			& 0.630 & 0.577 & 8.5\% \\
			\bottomrule
			
		\end{tabular}
	\end{adjustbox}
	\endgroup
\end{table*}
To rigorously evaluate the efficacy of the proposed EvoBrain framework, we conducted a comprehensive comparative analysis against state-of-the-art continual learning baselines under the Order 1 sequence. We benchmark against a diverse set of methodologies: MMD \cite{gretton2006kernel}, a distribution matching technique; EWC \cite{kirkpatrick2017overcoming} and LwF \cite{li2017learning}, which employ regularization and knowledge distillation to mitigate catastrophic forgetting; DER++ \cite{buzzega2020dark}, which combines episodic memory replay with stored logits and auxiliary self-supervision; AGEM \cite{chaudhry2018efficient}, which utilizes gradient episodic memory to constrain optimization updates; SDML \cite{wang2022learning}, a sequential domain meta-learning approach that employs a meta-optimizer with dynamic network freezing; CoUDA \cite{chen2025couda}, a recent continual domain adaptation framework; and AdaptFormer \cite{chen2022adaptformer}, a parameter-efficient fine-tuning strategy based on adapter modules.

The radar charts in Fig. \ref{fig:compared_result}(a)--(c) visualize the final retained performance across six heterogeneous EEG tasks under B-ACC, $\kappa$/AUC-PR, and WF1/AUROC. EvoBrain consistently forms a larger and more balanced performance envelope than competing methods, indicating stronger multi-task adaptability after the full continual learning sequence. In contrast, several baselines show sharp inward contractions on challenging datasets such as FACED and BCIC-2020-3, suggesting that conventional regularization, replay, or adapter-based strategies alone are insufficient to handle large cross-task discrepancies in EEG signals. Fig. \ref{fig:compared_result}(d) further summarizes the stability-plasticity trade-off using Average B-ACC and Average Forgetting. EvoBrain consistently lies in the favorable region with higher retained accuracy and lower forgetting, whereas most baselines exhibit a clearer compromise between adapting to later tasks and preserving earlier ones. This trend indicates that EvoBrain can maintain historical task competence while still accommodating newly arriving tasks, rather than improving plasticity at the cost of severe forgetting. An additional observation is that baseline methods tend to behave more competitively on the EEGMamba backbone than on several other backbones. This suggests that Mamba-based architectures may benefit from their selective state-space modeling, which can provide a relatively stable sequential representation and partially ease continual adaptation. Nevertheless, EvoBrain still achieves the most favorable overall trade-off, showing that architectural robustness alone is not sufficient to fully resolve heterogeneous cross-task continual learning. The task-wise behavior of AdaptFormer further highlights the difficulty of heterogeneous EEG continual learning. Since AdaptFormer freezes the backbone and only tunes lightweight adapters, it can perform reasonably well on simpler or representation-aligned tasks such as motor imagery (PhysioNet-MI) and mental disorder diagnosis (Mumtaz). However, it struggles on more complex tasks such as emotion recognition (FACED) and imagined speech (BCIC-2020-3), where the required representations may deviate substantially from the pre-training distribution. This contrast suggests that frozen-backbone adapters are not expressive enough for tasks involving high-order cognitive states and stronger distribution shifts. EvoBrain addresses this limitation by allowing both task-adaptive representation recalibration and response-level knowledge preservation during continual adaptation. Together, these mechanisms enable the model to accommodate the diverse representational demands of heterogeneous EEG tasks more effectively than frozen-backbone or conventional continual learning baselines. 
\begin{table*}[!ht]
	\centering
	\caption{Comparison of different adaptation paradigms using CBraMod as the shared EEG foundation backbone. 
		Single-task fine-tuning trains an isolated model for each task, multi-task fine-tuning jointly optimizes all tasks, whereas EvoBrain supports sequential task expansion through continual adaptation.}
	\label{tab:paradigm_comparison_cbramod}
	\resizebox{\textwidth}{!}{
		\begin{tabular}{lcccc|ccccccc}
			\toprule
			\multirow{2}{*}{Paradigm} 
			& \multicolumn{2}{c}{Paradigm Property}
			& \multicolumn{2}{c|}{Inference Cost}
			& \multicolumn{7}{c}{Performance} \\
			\cmidrule(lr){2-3}
			\cmidrule(lr){4-5}
			\cmidrule(lr){6-12}
			& New-Task Update
			& Evolvability
			& Backbones
			& Heads
			& ISRUC
			& PhysioNet-MI
			& FACED
			& BCIC2020-3
			& Mumtaz
			& BCIC-IV-2a
			& Avg. B-ACC / Retained \\
			\midrule
			Single-Task FT
			& Train new model
			& \xmark
			& $N\times$
			& $N\times$
			& 0.787±0.010
			& 0.642±0.009
			& 0.550±0.009
			& 0.537±0.011
			& 0.956±0.006
			& 0.514±0.007
			& 0.664±0.009 / -- \\
			
			Multi-Task FT
			& Retrain on all tasks
			& \xmark
			& $1\times$
			& $N\times$
			& 0.759±0.007
			& 0.608±0.011
			& 0.545±0.020
			& 0.421±0.020
			& 0.889±0.007
			& 0.521±0.013
			& 0.624±0.003 / 94.0\% \\
			
			EvoBrain
			& Incremental update
			& \cmark
			& $1\times$
			& $N\times$
			& 0.729±0.004
			& 0.546±0.007
			& 0.475±0.004
			& 0.482±0.007
			& 0.876±0.010
			& 0.527±0.008
			& 0.606±0.002 / 91.3\% \\
			\bottomrule
		\end{tabular}
	}
\end{table*}

\subsection{Ablation Study}
To elucidate the individual contributions of EvoBrain, we conduct a systematic ablation analysis across four EEG foundation models, as reported in Tab. \ref{tab:ablation_study}. Overall, the complete framework achieves the strongest or tied-strongest retained Average B-ACC while maintaining consistently low RFR across different backbones, indicating that its performance gain comes from both effective task adaptation and stable knowledge retention. Removing Neuro-Spectral Task Normalization (NSN) leads to clear degradation in retained performance for most backbones, especially for CBraMod, CSBrain, and DeeperBrain. This suggests that heterogeneous EEG tasks cannot be handled by replay or distillation alone; the model first needs a task-aware normalization mechanism to reduce cross-task representation mismatch. The finer-grained ablations further show that Prototype-guided Temporal Normalization (PTN) and Spectral Residual Modulation (SRM) play complementary roles: PTN stabilizes task-level temporal statistics, whereas SRM recalibrates task-related spectral responses, which is particularly important when the task stream contains paradigms with distinct frequency signatures. The ablation of Response-Affinity Distillation (RAD) further demonstrates the importance of response-level regularization for continual EEG adaptation. Removing RAD generally reduces the final Average B-ACC and increases forgetting, indicating that normalizing new-task representations alone is insufficient to preserve historical task competence. Among its two subcomponents, Response-Geometry Distillation (RGD) mainly contributes to old-task stability by preserving the relational structure of replayed responses, while Spectral-Affinity Alignment (SAA) improves cross-task transfer by aligning response statistics only between tasks with compatible spectral signatures. This design avoids indiscriminate feature alignment across unrelated EEG paradigms and helps balance stability and plasticity. Notably, EEGMamba shows smaller variation across several ablation variants, suggesting that Mamba-based selective state-space modeling may provide a relatively robust representation basis for continual adaptation. Nevertheless, the complete EvoBrain still offers the most reliable overall trade-off, confirming that NSN and RAD provide complementary benefits beyond the intrinsic robustness of individual backbone architectures.

\begin{figure*}[!tb]
	\centering
	\includegraphics[width=1.0\textwidth]{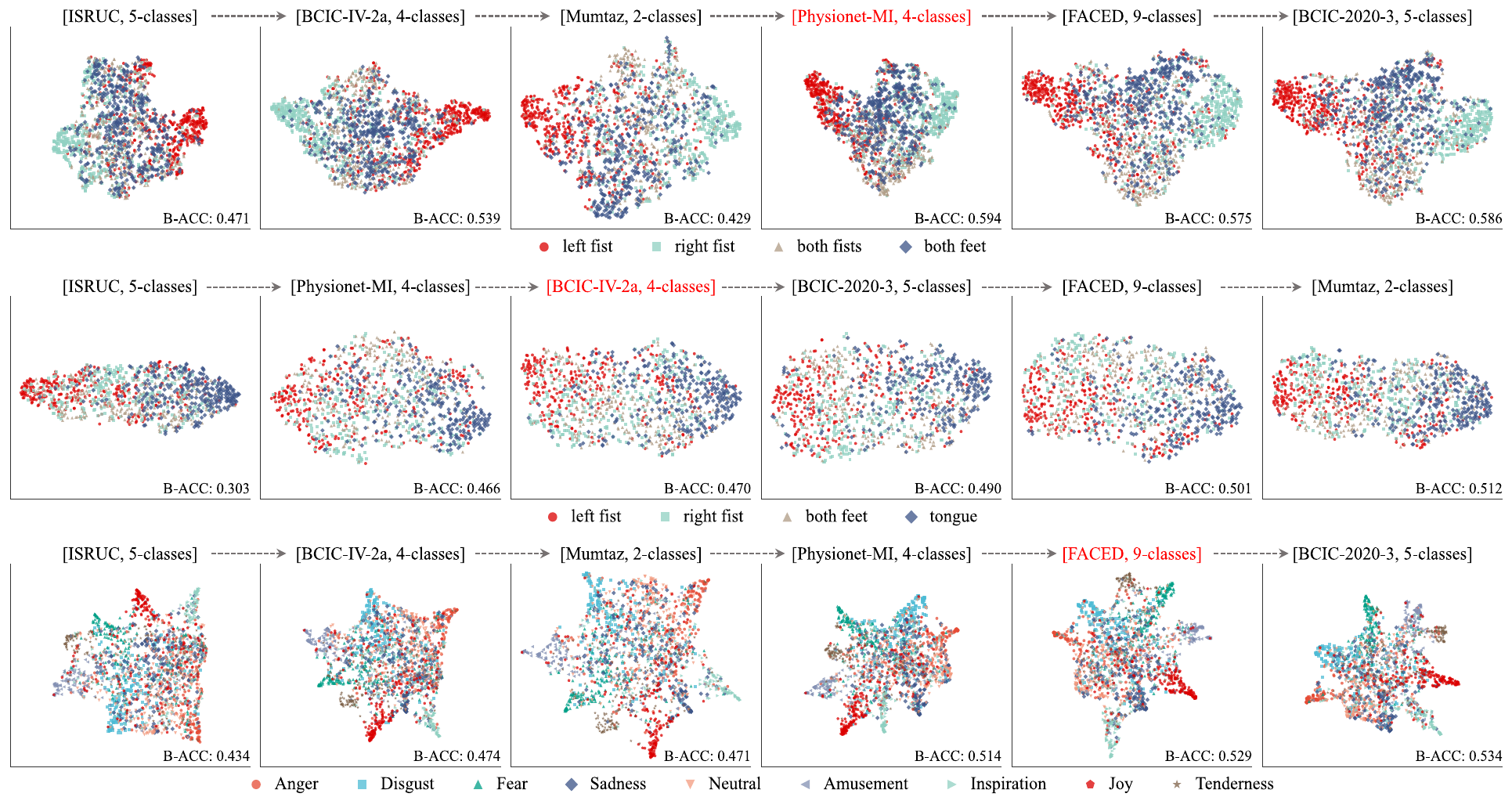}
	\caption{Longitudinal evolution of latent representations and classification performance throughout the cross-task continual learning process. The first and second rows visualize the Physionet-MI and BCIC-IV-2a datasets using CBraMod, respectively, while the third row visualizes the FACED dataset using DeeperBrain. The stages marked in red denote the onset of training for the corresponding target tasks. The reported Balanced Accuracy (B-ACC) scores show that related tasks can induce positive transfer before explicit target-task adaptation, as reflected by the early improvement between the two motor imagery datasets (Physionet-MI: $0.471 \rightarrow 0.539$ after BCIC-IV-2a adaptation; BCIC-IV-2a: $0.303 \rightarrow 0.466$ after Physionet-MI adaptation). After target-task adaptation, the final B-ACC remains high for Physionet-MI, BCIC-IV-2a, and FACED ($0.586$, $0.512$, and $0.534$).}
	\label{fig:vis}
\end{figure*}
\subsection{Continual Expandability versus Static Fine-Tuning}
Table~\ref{tab:paradigm_comparison_cbramod} presents a paradigm-level comparison of single-task fine-tuning, multi-task fine-tuning, and EvoBrain using CBraMod as the shared EEG foundation backbone. Here, "Retained" denotes the percentage of average B-ACC preserved relative to single-task fine-tuning, which serves as a task-isolated upper-bound reference by independently optimizing a dedicated backbone for each task. Although this strategy achieves the highest average performance, it requires full fine-tuning and storage of a separate backbone for every task, causing training and deployment costs to increase linearly as the number of tasks grows. Multi-task fine-tuning achieves slightly higher average B-ACC than EvoBrain with only one shared backbone, but it assumes that all task datasets are jointly accessible during training. When a new task arrives, the model must be retrained on the union of all previous and new task data, resulting in substantial data-access and computational costs. In contrast, EvoBrain operates under a sequential task stream, where each new task is incorporated through incremental adaptation without full retraining. With only one shared backbone, EvoBrain retains 91.3\% of the single-task performance while supporting continual task expansion, demonstrating better scalability and long-term adaptability for realistic scenarios in which new EEG tasks emerge over time.

\subsection{Visualization of Cross-Task Representation Evolution}
To further examine how EvoBrain reorganizes neural representations during continual adaptation, we visualize the longitudinal evolution of target-task latent spaces and their corresponding B-ACC scores in Fig. \ref{fig:vis}. This visualization tracks how the representation of a selected target task changes as the model sequentially adapts to other heterogeneous EEG tasks. Specifically, the first and second rows analyze two motor imagery tasks, PhysioNet-MI and BCIC-IV-2a, using CBraMod, while the third row analyzes the emotion recognition task FACED using DeeperBrain. This setup allows us to inspect both cross-paradigm stability and within-paradigm transfer throughout the continual learning process. 

The first observation is that related tasks suggest possible positive forward-transfer behavior before explicit target-task training. In the first row, PhysioNet-MI has not yet been introduced when the model adapts to BCIC-IV-2a, yet its B-ACC increases from 0.471 to 0.539. A similar pattern appears in the second row: BCIC-IV-2a improves from 0.303 to 0.466 after adapting to PhysioNet-MI. These reciprocal gains suggest that EvoBrain can extract transferable paradigm-level representations and improve target-task separability even before direct supervision on the target dataset. More importantly, this effect is not limited to closely related motor imagery tasks. As shown in the third row, FACED also improves from 0.434 to 0.514 after the model has sequentially adapted to four datasets that are not directly associated with emotion recognition. This broader improvement indicates that EvoBrain may accumulate cross-task representational structure, providing a stronger representational basis for subsequent target-task adaptation. The second observation is that target-task representations remain recoverable and can continue to improve after their own adaptation stage. For PhysioNet-MI, the trajectory is not strictly monotonic, dropping to 0.429 after Mumtaz, but it recovers strongly when the target task is trained and remains high at the end of the sequence, reaching 0.586 after all tasks. For BCIC-IV-2a, the representation continues to improve after its own training stage, rising from 0.470 to a final B-ACC of 0.512. The FACED trajectory also shows continual refinement: its B-ACC further increases from 0.529 to 0.534 after the subsequent BCIC-2020-3 task. These trends indicate that later task adaptation does not necessarily overwrite previous knowledge; instead, compatible subsequent tasks can further reorganize the latent space and improve target-task discriminability. Overall, this visualization offers qualitative support for EvoBrain’s ability to maintain and progressively adjust task representations throughout continual adaptation across heterogeneous EEG streams.


\subsection{Visualization of Spectral-Modulation Patterns}
\begin{figure*}[!t]
	\centering
	\includegraphics[width=1.0\textwidth]{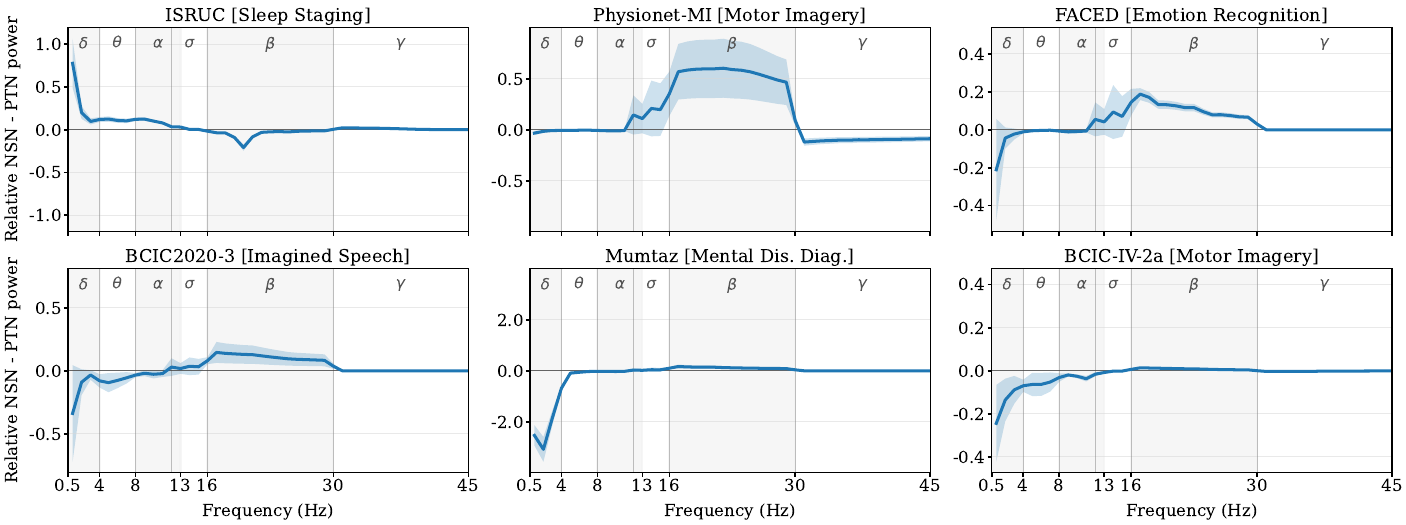}
	\caption{
		Visualization of NSN-induced relative linear spectral modulation over five repeated CBraMod runs under Order 1.
		For each task, we use the final student model after continual learning and compute the frequency-wise power change induced by the Spectral Residual Modulation (SRM) component of NSN. The y-axis reports \(\Delta_{\mathrm{rel}}(f)\), as defined in Eq.~\eqref{eq_relative_power_diff}, which measures the normalized frequency-wise power change induced by NSN relative to the temporally normalized representation.
		The solid line denotes the mean over five repeated CBraMod runs, and the shaded region denotes the 95\% confidence interval.
		Vertical bands indicate canonical EEG frequency ranges: \(\delta\), \(\theta\), \(\alpha\), \(\sigma\), \(\beta\), and \(\gamma\).
	}
	\label{fig:nsn_vis}
\end{figure*}

We further visualize the frequency-wise effect of Neuro-Spectral Task Normalization (NSN) by examining how its Spectral Residual Modulation (SRM) component reshapes task-specific EEG representations. The visualization is conducted on CBraMod under the Order 1 continual learning stream. For each task, we use the final student model and compute the relative linear spectral modulation \(\Delta_{\mathrm{rel}}(f)\) defined in Eq.~\eqref{eq_relative_power_diff}. We report the mean curve over five repeated CBraMod runs, with the shaded region indicating the 95\% confidence interval. The results suggest that SRM learns task-conditioned spectral recalibrations that are qualitatively consistent with canonical EEG rhythm priors. For sleep staging, SRM emphasizes the low-frequency \(\delta\) range, aligning with the relevance of slow-wave activity in sleep analysis. For motor imagery, PhysioNet-MI shows a clear \(\beta\)-band enhancement, whereas BCIC-IV-2a mainly suppresses low-frequency components with only mild \(\beta\)-band amplification. For emotion recognition, SRM also increases the \(\beta\) range, suggesting that it tends to emphasize potentially discriminative spectral components for affective EEG representations. Overall, these patterns indicate that NSN performs task-conditioned residual spectral adjustment rather than uniform normalization. This analysis provides a model-level interpretation of the learned SRM behavior: the observed correspondence with EEG rhythm priors offers qualitative support for the interpretability of NSN, rather than direct neurophysiological evidence.

\subsection{Visualization of Relation-Structure Preservation.}
\begin{figure*}[!t]
	\centering
	\includegraphics[width=1.0\textwidth]{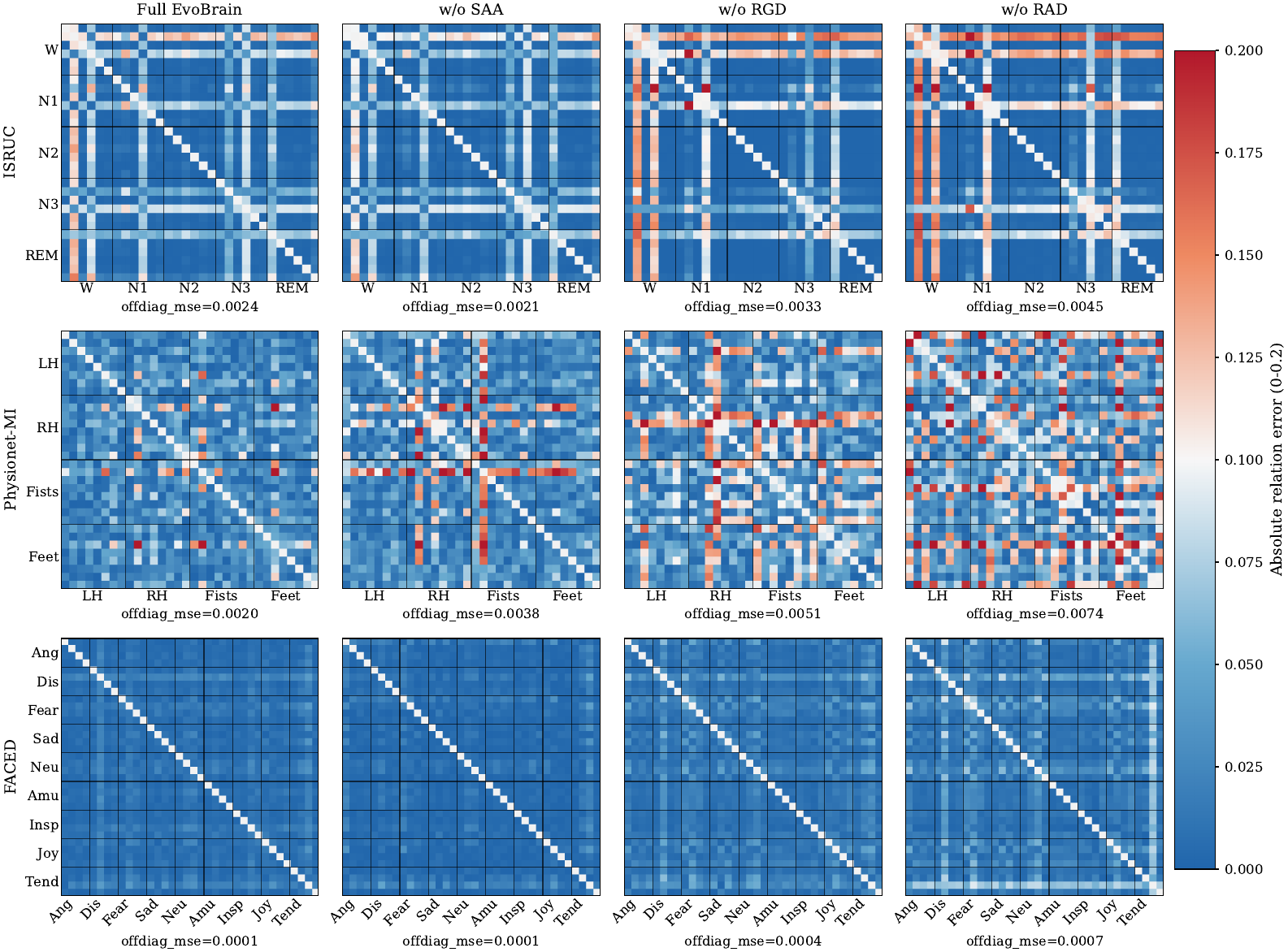}
	\caption{
		Visualization of old-task relation preservation under RAD. We use CBraMod under the Order-1 task stream and select the first three old tasks, ISRUC, PhysioNet-MI, and FACED. For each task $T_i$ ($i \in \{1,2,3\}$), $\mathcal{M}_i$ denotes the frozen teacher after adapting to $T_i$, and $\mathcal{M}_\mathcal{N}$ denotes the final student after the full six-task sequence. Following Eqs.~\eqref{eq19}--\eqref{eq21}, we compute sample-wise cosine relation matrices from pooled intermediate responses of the teacher and student on the same replay samples, and visualize their absolute difference $|R_{\mathrm{student}} - R_{\mathrm{teacher}}|$. The value below each subplot reports the off-diagonal mean squared error (\texttt{offdiag\_mse}), where smaller values and colder colors indicate better preservation of old-task response geometry. Full EvoBrain and w/o SAA consistently show lower relation errors than w/o RGD and w/o RAD, especially on ISRUC and PhysioNet-MI, demonstrating that RGD is the key component for stabilizing old-task relational structures, while SAA provides complementary response-statistics alignment.
	}
	\label{fig:heatmap}
\end{figure*}
We further visualize whether RAD preserves the internal response geometry of previously learned tasks by comparing the teacher--student relation matrices on replayed old-task samples. Specifically, for the first three tasks in the Order-1 stream, i.e., ISRUC, PhysioNet-MI, and FACED, we use the corresponding teacher checkpoints $\mathcal{M}_i$ and the final student model $\mathcal{M}_6$. Specifically, we extract the intermediate responses from the teacher and student and compute the sample-wise cosine relational matrices as defined in Eq. \ref{eq19} and Eq. \ref{eq20}. We then visualize the absolute difference $|R_{\mathrm{student}} - R_{\mathrm{teacher}}|$, where lower error indicates that the final student better preserves the old-task response geometry encoded by the frozen teacher. The reported \texttt{offdiag\_mse} is computed only on the off-diagonal entries, exactly following the RGD objective in Eq. \ref{eq21}. Figure~7 visualizes old-task relation preservation under RAD. Full EvoBrain consistently keeps relation errors low across the first three old tasks, while the variant that removes SAA remains close to the full model, suggesting that removing SAA alone does not substantially disrupt old-task relational geometry. In contrast, removing RGD produces stronger error bands and larger off-diagonal MSE across all three datasets, with the most evident degradation on PhysioNet-MI, where the error increases from 0.0020 to 0.0051. The consistent trend observed on ISRUC and FACED further identifies RGD as the key factor for preserving old-task relational structures. This result shows that merely replaying historical samples is insufficient, since replay provides access to old-task data but does not explicitly constrain the geometry of their responses. RGD addresses this limitation by aligning the student’s replay-sample relation matrix with that of the corresponding teacher, thereby suppressing inter-sample relational drift during continual adaptation. 

\section{Discussion}
\subsection{Implications}
The present results suggest that cross-task continual adaptation is a promising post-training paradigm for EEG foundation models. Unlike task-isolated fine-tuning, which repeatedly optimizes and stores separate backbones for different downstream tasks, EvoBrain enables a single backbone to evolve across sequentially arriving heterogeneous BCI tasks. Its ability to retain 91.3\% of the single-task fine-tuning performance indicates that scalable one-for-all EEG decoding requires not only larger pretraining corpora, but also post-training mechanisms that support continual task expansion without restarting adaptation from the pretrained model. The effectiveness of NSN further shows that continual EEG adaptation should be treated as both a distributional and neuro-spectral alignment problem. By combining prototype-guided temporal normalization with task-conditioned spectral residual modulation, NSN stabilizes cross-task feature statistics while preserving task-specific frequency responses, moving beyond generic feature normalization toward EEG-specific spectral recalibration. Meanwhile, RAD demonstrates that stability in heterogeneous task streams depends on preserving relational response structure and enabling selective transfer, rather than merely constraining individual predictions or parameters. Together, these findings position EvoBrain as a neuro-spectrally grounded continual adaptation framework, where plasticity is achieved through task-adaptive spectral recalibration and stability is maintained through affinity-aware response preservation.

\subsection{Limitations}

Several directions could further extend EvoBrain toward more open BCI applications. The current study adopts a controlled task-incremental protocol with known task boundaries and task-specific heads, which enables systematic evaluation of cross-task continual adaptation. Future work may consider more flexible online settings with ambiguous task identity, delayed labels, dynamic head selection, or unified output spaces. In addition, although replay memory effectively improves stability, privacy-aware alternatives such as federated continual learning, prototype replay, and generative replay could further facilitate deployment in clinical or multi-institutional environments. Finally, the spectral modulation analysis provides model-level interpretability, and future studies with controlled paradigms and subject-level validation may further strengthen its neurophysiological grounding.
\section{Conclusion}
In this paper, we presented EvoBrain, a cross-task continual learning framework for adapting EEG foundation models toward one-for-all brain decoding. Instead of relying on task-isolated fine-tuning, EvoBrain enables a single pretrained backbone to incrementally incorporate heterogeneous BCI tasks while retaining previously acquired decoding capability. The framework is built on two complementary components: Neuro-Spectral Task Normalization (NSN), which mitigates cross-task distribution shifts through prototype-guided temporal normalization and task-conditioned spectral residual modulation; and Response-Affinity Distillation (RAD), which preserves old-task response geometry and promotes selective transfer according to spectral affinity. Extensive experiments on six EEG tasks and four foundation backbones, including Transformer-based and Mamba-based architectures, show that EvoBrain consistently improves the stability-plasticity trade-off over representative continual learning baselines and supports sequential task expansion with a single evolving backbone. Additional analyses of representation evolution, spectral modulation, and response-structure preservation further suggest that EvoBrain learns transferable cross-task representations while maintaining task-specific neuro-spectral characteristics. To the best of our knowledge, this work provides the first systematic study of task-level continual learning for EEG foundation models, offering a practical post-training direction for scalable and general-purpose neural decoding systems.

\begin{footnotesize}
	\bibliographystyle{IEEEtranN}
	\bibliography{references}
\end{footnotesize}

\section{Biography Section}
\begin{IEEEbiography}[{\includegraphics[width=1in,height=1.25in,clip,keepaspectratio]{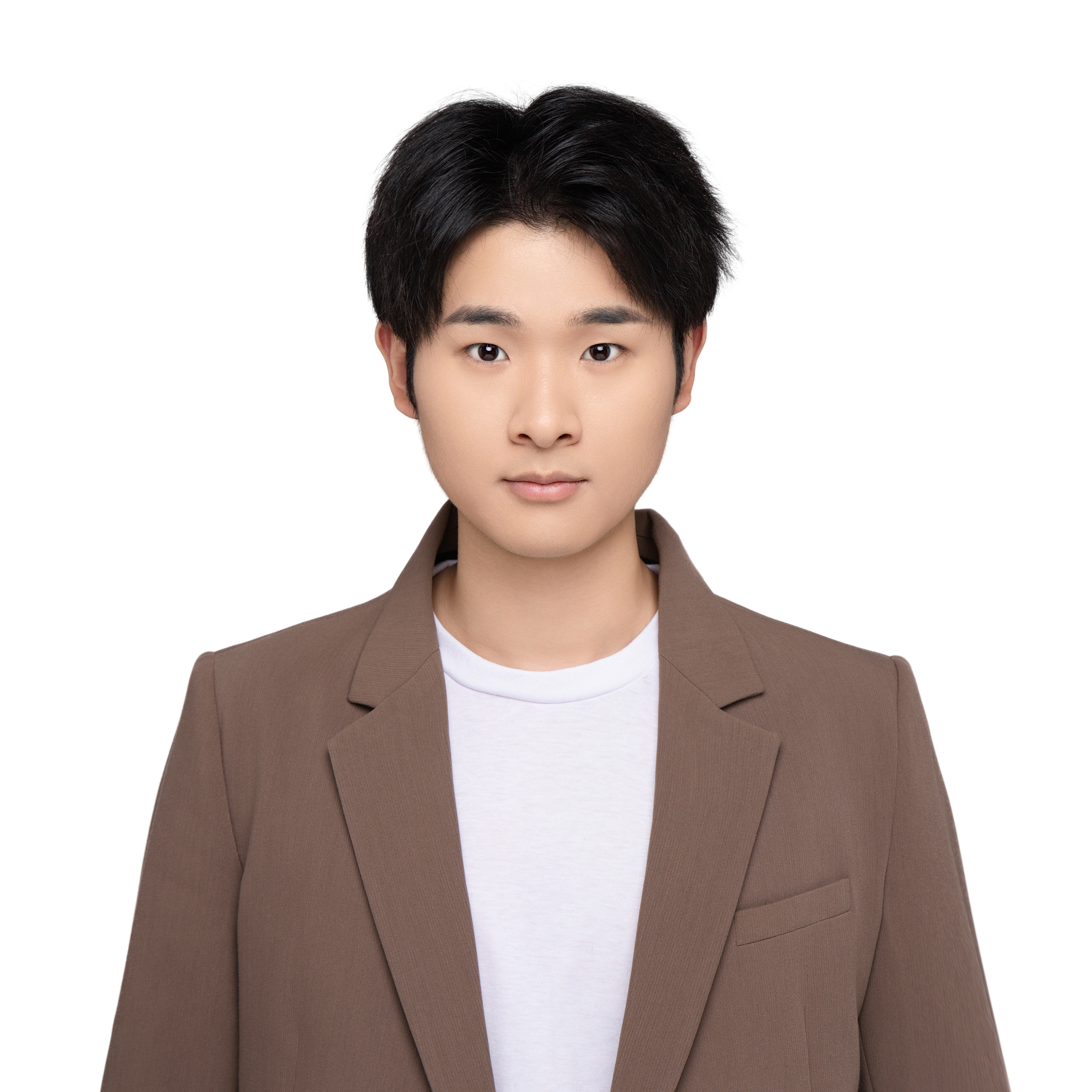}}]{Yangxuan Zhou}
	received the B.S. degree in Electronic Information from Shanghai University, Shanghai, China, in 2022.
	
	He is currently working toward the Ph.D. degree of Computer Science and Technology with Zhejiang University, Hangzhou, China. His research interests include EEG decoding, brain-computer interfaces, deep learning, and artificial intelligence.
\end{IEEEbiography}
\begin{IEEEbiography}[{\includegraphics[width=1in,height=1.25in,clip,keepaspectratio]{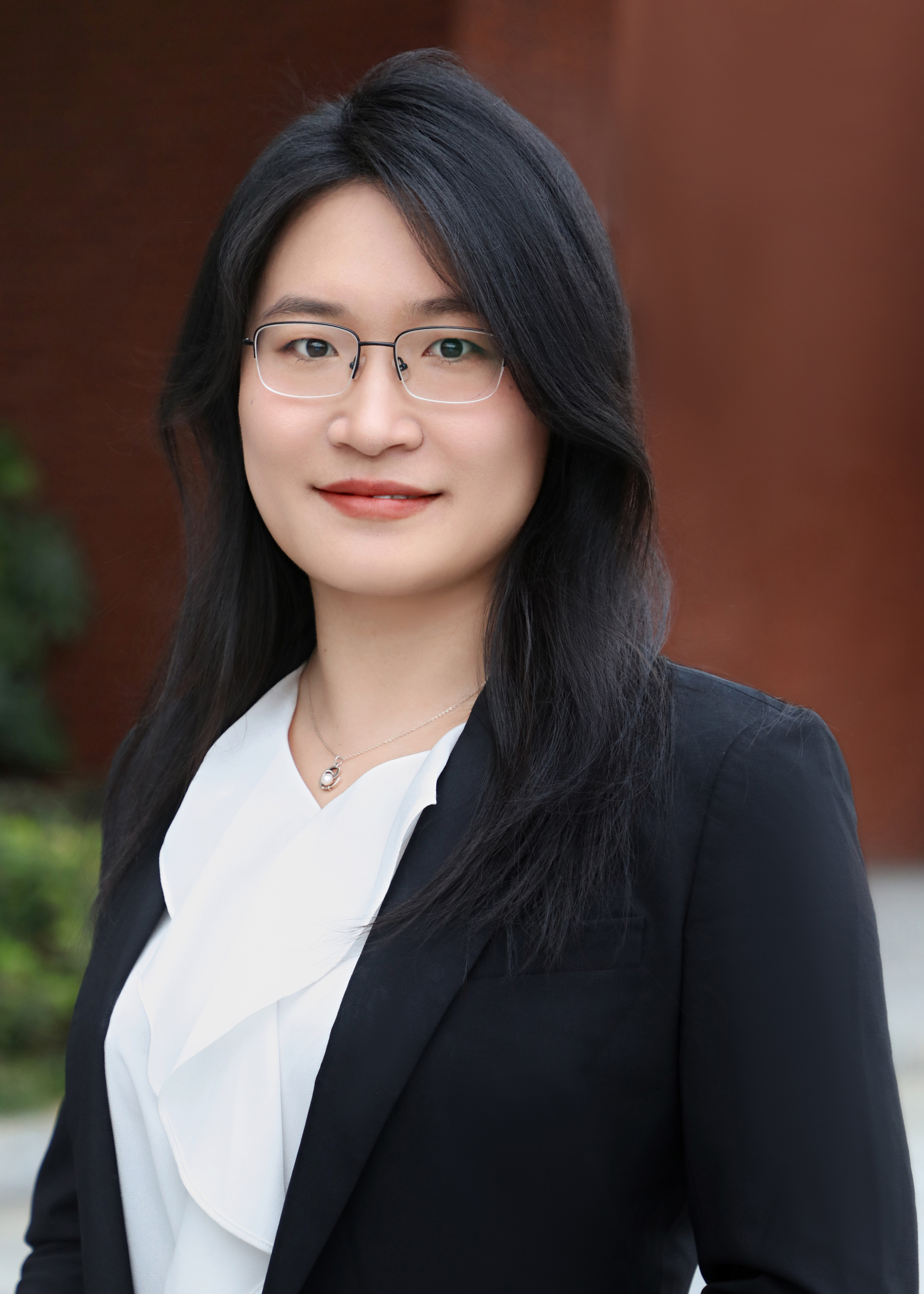}}]{Sha Zhao}
	received the Ph.D. degree from Zhejiang
	University, Hangzhou, China, in 2017. 
	
	She is currently a research professor with the
	College of Computer Science and Technology, Zhejiang University. She visited the Human-Computer
	Interaction Institute, Carnegie Mellon University,
	Pittsburgh, PA, USA, as a visiting Ph.D. student
	from 2015 to 2016. Her research interests include
	brain-machine interfaces, data mining and machine
	learning. Dr. Zhao received the Best Paper Award of
	ACM UbiComp’16.
\end{IEEEbiography}
\begin{IEEEbiography}[{\includegraphics[width=1in,height=1.25in,clip,keepaspectratio]{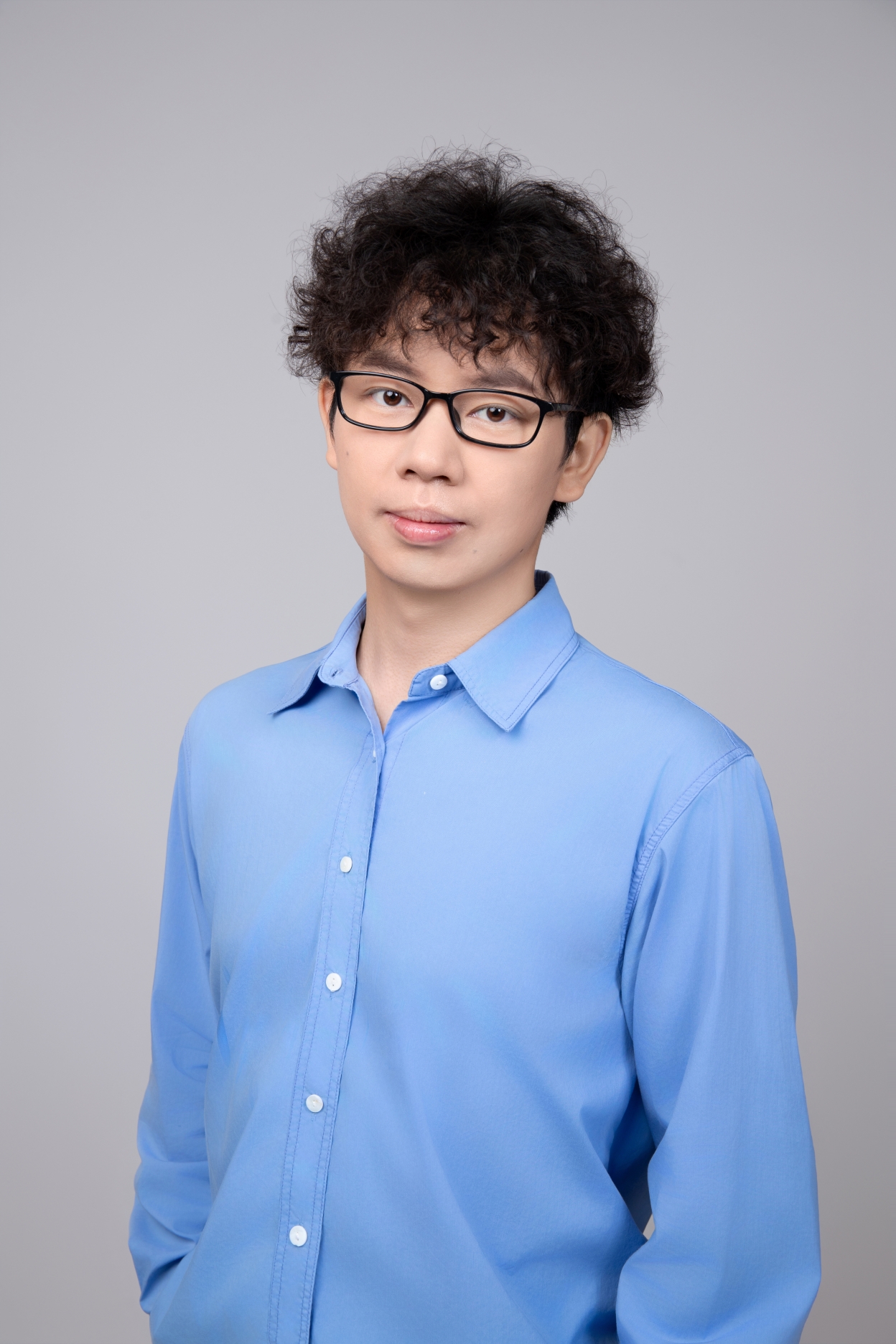}}]{Jiquan Wang}
	received the Ph.D. degree in Computer Science and Technology from Zhejiang University, Hangzhou, China, in 2025.
	
	He is currently a research fellow at the State Key Laboratory of Brain-Machine Intelligence, Zhejiang University. His research interests include EEG decoding, brain-computer interfaces and artificial intelligence. He received the 2025 ACM Hangzhou Outstanding Doctoral Dissertation Award.
\end{IEEEbiography}
\begin{IEEEbiography}[{\includegraphics[width=1in,height=1.25in,clip,keepaspectratio]{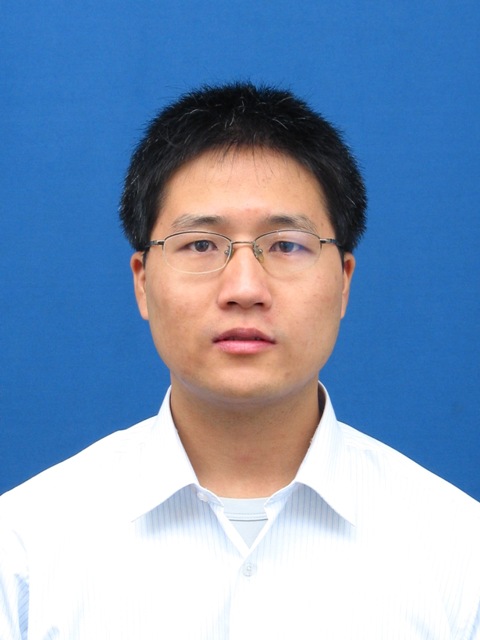}}]{Shijian Li}
	received the Ph.D. degree from Zhejiang
	University, Hangzhou, China, in 2006.
	
	In 2010, he was a Visiting Scholar with the
	Institute Telecom SudParis, Évry, France. He is currently with the College of Computer Science and Technology, Zhejiang University. He has published over 40 papers. His research interests include sensor networks, ubiquitous computing, and social computing.
	Dr. Li serves as an Editor for the International Journal of Distributed Sensor Networks and as a reviewer or the PC member of over ten conferences.
\end{IEEEbiography}
\begin{IEEEbiography}[{\includegraphics[width=1in,height=1.25in,clip,keepaspectratio]{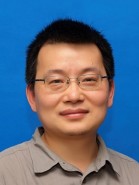}}]{Gang Pan} 
	(Senior Member, IEEE) received the B.Eng. and
	Ph.D. degrees from Zhejiang University, Hangzhou, China, in 1998 and 2004, respectively.
	
	He is currently a Professor with the Department of Computer Science, and the Deputy Director of the State Key Lab of CAD\&CG, Zhejiang University. From 2007 to 2008, he was a Visiting Scholar with
	the University of California, Los Angeles, CA, USA. His current interests include brain-inspired computing, brain-machine interfaces, artificial intelligence and pervasive computing.
\end{IEEEbiography}

\end{document}